%% file: main_arxiv.tex
\documentclass[10pt,twocolumn,letterpaper]{article}

\usepackage{iccv}
\usepackage{times}
\usepackage{epsfig}
\usepackage{graphicx}
\usepackage{amsmath}
\usepackage{amssymb}

\usepackage{microtype}
\usepackage{graphicx}
\usepackage{caption}
\usepackage{subcaption}
\usepackage{booktabs} 
\usepackage[table]{xcolor}

\usepackage{multirow}
\usepackage{algorithm}
\usepackage{algorithmic}
\usepackage{xspace}
\usepackage{bm}
\usepackage{fancyhdr}

\usepackage[pagebackref=true,breaklinks=true,letterpaper=true,colorlinks,bookmarks=false]{hyperref}

\usepackage{cleveref}

\iccvfinalcopy 

\def\iccvPaperID{****} 

\newcommand\blfootnote[1]{%
  \begingroup
  \renewcommand\thefootnote{}\footnote{#1}%
  \addtocounter{footnote}{-1}%
  \endgroup
}

\begin{document}

\title{Q-Diffusion: Quantizing Diffusion Models}

\author{Xiuyu Li\\
UC Berkeley\\
\and
Yijiang Liu\\
Nanjing University\\
\and
Long Lian\\
UC Berkeley\\
\and
Huanrui Yang\\
UC Berkeley\\
\and
Zhen Dong\\
UC Berkeley\\
\and
Daniel Kang\\
UIUC 
\and
Shanghang Zhang\\
Peking University\\
\and
Kurt Keutzer\\
UC Berkeley\\
}

\maketitle


\makeatletter
\def\blfootnote{\gdef\@thefnmark{}\@footnotetext}
\makeatother

\blfootnote{\!\!\!\!\!\!\!\!Preprint. Under review.}

\input{macro.tex}

\input{0abstract.tex}

\input{1introduction.tex}
\input{2background.tex}
\input{3method.tex}

\input{4experiment.tex}
\input{6conclusion.tex}

{\small
\bibliographystyle{ieee_fullname}
\bibliography{ref}
}

\newpage
\appendix
\onecolumn
\input{appendix.tex}

\end{document}

%% file: macro.tex
\newcommand{\sect}[1]{Section~\ref{#1}}
\newcommand{\ssect}[1]{\S~\ref{#1}}
\newcommand{\append}[1]{Appendix~\ref{#1}}
\newcommand{\eqn}[1]{Equation~\ref{#1}}
\newcommand{\fig}[1]{Figure~\ref{#1}}
\newcommand{\tbl}[1]{Table~\ref{#1}}
\newcommand{\algo}[1]{Algorithm~\ref{#1}}
\newcommand{\name}{Q-Diffusion\xspace}
\newcommand{\cifar}{CIFAR-10\xspace}
\newcommand{\bed}{LSUN-Bedrooms\xspace}
\newcommand{\church}{LSUN-Churches\xspace}
\newcommand{\sd}{Stable Diffusion\xspace}
\newcommand{\myparagraph}[1]{\noindent \textbf{#1}}
\newcommand{\na}{---}
\newcommand{\ourcell}{\cellcolor[rgb]{1,0.808,0.576}}

\renewcommand{\vec}[1]{\boldsymbol{\mathbf{#1}}}

\newcommand{\todo}[1]{\textcolor{red}{[TODO: #1]}}
\newcommand{\tony}[1]{\textcolor{blue}{[Tony: #1]}}
\newcommand{\xiuyu}[1]{\textcolor{magenta}{[Xiuyu: #1]}}
\newcommand{\yijiang}[1]{\textcolor{cyan}{[Yijiang: #1]}}
\newcommand{\huanrui}[1]{\textcolor{yellow}{[Huanrui: #1]}}
\newcommand{\zhen}[1]{\textcolor{orange}{[Zhen: #1]}}
\newcommand{\update}[1]{\textcolor{blue}{#1}}

%% file: 0abstract.tex
\begin{abstract}
   Diffusion models have achieved great success in image synthesis through iterative noise estimation using deep neural networks. However, the slow inference, high memory consumption, and computation intensity of the noise estimation model hinder the efficient adoption of diffusion models. Although post-training quantization~(PTQ) is considered a go-to compression method for other tasks, it does not work out-of-the-box on diffusion models. We propose a novel PTQ method specifically tailored towards the unique multi-timestep pipeline and model architecture of the diffusion models, which compresses the noise estimation network to accelerate the generation process.
   We identify the key difficulty of diffusion model quantization as the changing output distributions of noise estimation networks over multiple time steps and the bimodal activation distribution of the shortcut layers within the noise estimation network.
   We tackle these challenges with timestep-aware calibration and split shortcut quantization in this work.
   Experimental results show that our proposed method is able to quantize full-precision unconditional diffusion models into 4-bit while maintaining comparable performance (small FID change of at most 2.34 compared to $>$100 for traditional PTQ) in a training-free manner. Our approach can also be applied to text-guided image generation, where we can run stable diffusion in 4-bit weights with high generation quality for the first time.
\end{abstract}
\vspace{-10pt}

%% file: 1introduction.tex
\section{Introduction}
\label{sec:introduction}

Diffusion models have shown great success in generating images with both high diversity and high fidelity \cite{song2019generative, ho2020denoising, song2020score, song2020denoising,dhariwal2021diffusion, ramesh2022hierarchical, saharia2022photorealistic, Rombach2021HighResolutionIS}. Recent work \cite{karras2020analyzing, karras2020training} has demonstrated superior performance than state-of-the-art GAN models, which suffer from unstable training. As a class of flexible generative models, diffusion models demonstrate their power in various applications such as image super-resolution \cite{Saharia2021ImageSV, Li2021SRDiffSI}, inpainting \cite{song2020score}, shape generation \cite{Cai2020LearningGF}, graph generation \cite{Niu2020PermutationIG}, image-to-image translation \cite{Sasaki2021UNITDDPMUI}, and molecular conformation generation \cite{Xu2022GeoDiffAG}.

However, the generation process for diffusion models can be slow due to the need for an iterative noise estimation of 50 to 1,000 time steps \cite{ho2020denoising, song2020denoising} using complex neural networks. 
While previous state-of-the-art approaches (e.g., GANs) are able to generate multiple images in under 1 second, it normally takes several seconds for a diffusion model to sample a single image. Consequently, speeding up the image generation process becomes an important step toward broadening the applications of diffusion models. Previous work has been solving this problem by finding shorter, more effective sampling trajectories \cite{song2020denoising, nichol2021improved, Salimans2022ProgressiveDF, liu2022pseudo, Bao2022AnalyticDPMAA, Lu2022DPMSolverAF}, which reduces the number of steps in the denoising process. 
However, they have largely ignored another important factor: the noise estimation model used in each iteration itself is compute- and memory-intensive. This is an orthogonal factor to the repetitive sampling, which not only slows down the inference speed of diffusion models, but also poses crucial challenges in terms of high memory footprints.

\begin{figure*}[h]
\centering
\vspace{-5pt}
\includegraphics[width=0.9\linewidth]{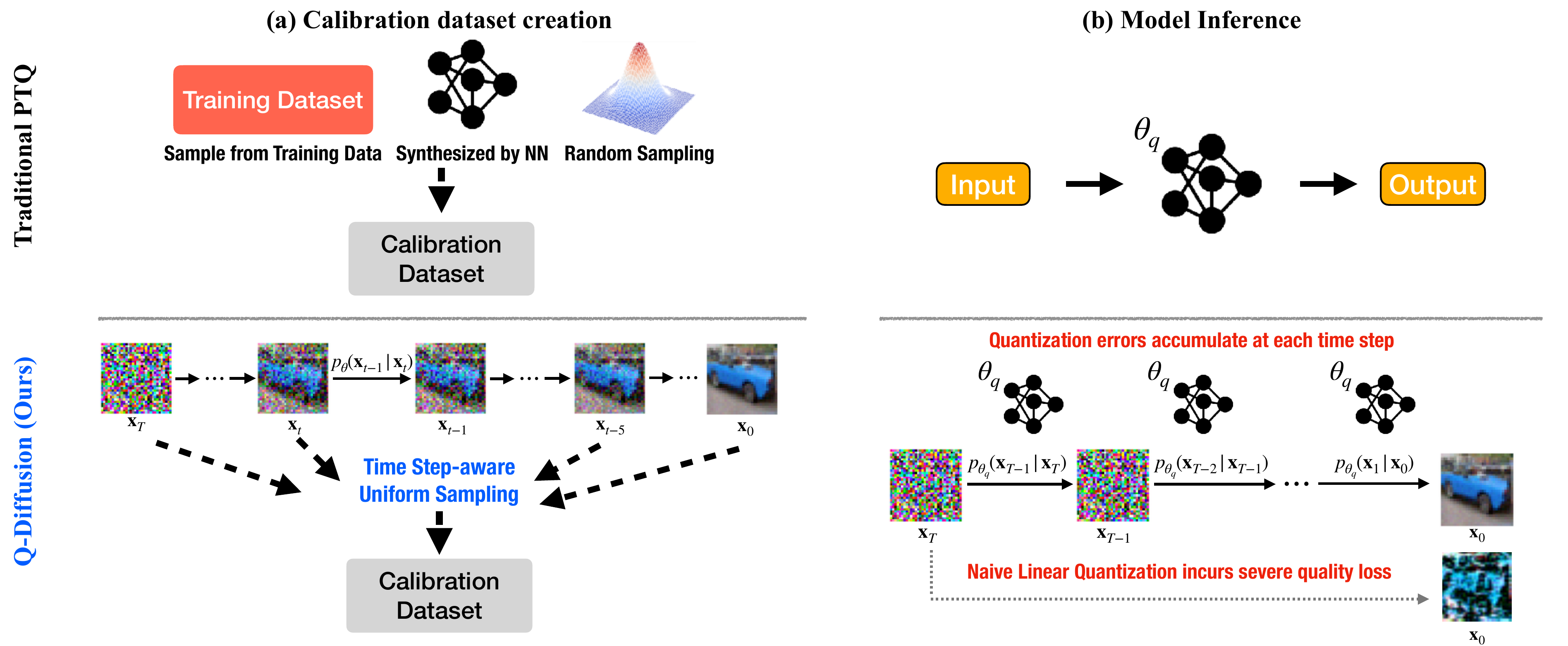}
\caption{Conventional PTQ scenarios and \name differ in (a) calibration dataset creation and (b) model inference workflow. Traditional PTQ approaches sample data randomly \cite{Guo2022SQuantOD}, synthesize with statistics in model layers \cite{Cai2020ZeroQAN}, or draw from the training set to create calibration dataset \cite{Nagel2020UpOD, li2021brecq}, which either contains inconsistency with real inputs during the inference time or are not data-free. In contrast, \name constructed calibration datasets with inputs that are an accurate reflection of data seen during the production in a data-free manner. Traditional PTQ inference only needs to go through the quantized model $\theta_q$ one time, while \name needs to address the accumulated quantization errors in the multi-time step inference.}
\label{hook}
\vspace{-0.3cm}
\end{figure*}

This work explores the compression of the noise estimation model used in the diffusion model to accelerate the denoising of all time steps. Specifically, we propose exploring post-training quantization (PTQ) on the diffusion model. PTQ has already been well studied in other learning domains like classification and object detection~\cite{Cai2020ZeroQAN, Bhalgat2020LSQIL, li2021brecq, Guo2022SQuantOD, liu2022noisyquant}, and has been considered a go-to compression method given its minimal requirement for training data and the straightforward deployment on real hardware devices.
However, the iterative computation process of the diffusion model and the model architecture of the noise estimation network brings unique challenges to the PTQ of diffusion models. PTQ4DM~\cite{shang2022posttraining} presents an inaugural application of PTQ to compress diffusion models down to 8-bit, but it primarily focuses on smaller datasets and lower resolutions.

Our work, evolving concurrently with~\cite{shang2022posttraining}, offers a comprehensive analysis of the novel challenges of performing PTQ on diffusion models. Specifically, as visualized in~\fig{hook}(a), we discover that the output distribution of the noise estimation network at each time step can be largely different, and naively applying previous PTQ calibration methods with an arbitrary time step leads to poor performance. Furthermore, as illustrated in~\fig{hook}(b), the iterative inference of the noise estimation network leads to an accumulation of quantization error, which poses higher demands on designing novel quantization schemes and calibration objectives for the noise estimation network. 

To address these challenges, we propose \textbf{Q-Diffusion}, a PTQ solution to compress the cumbersome noise estimation network in diffusion models in a data-free manner, while maintaining comparable performance to the full precision counterparts.
We propose a time step-aware calibration data sampling mechanism from the pretrained diffusion model, which represents the activation distribution of all time steps. We further tailor the design of the calibration objective and the weight and activation quantizer to the commonly used noise estimation model architecture to reduce quantization error.
We perform thorough ablation studies to verify our design choices, and demonstrate good generation results with diffusion models quantized to only 4 bits.

In summary, our contributions are:
\begin{enumerate}
\item We propose \textbf{Q-Diffusion}, a data-free PTQ solution for the noise estimation network in diffusion models.
\item We identify the novel challenge of performing PTQ on diffusion models as the activation distribution diversity and the quantization error accumulation across time steps via a thorough analysis.
\item We propose time step-aware calibration data sampling to improve calibration quality, and propose a specialized quantizer for the noise estimation network.
\item Extensive results show Q-Diffusion enables W4A8 PTQ for both pixel-space and latent-space unconditional diffusion models with an FID increment of only $0.39 {\text-} 2.34$ over full precision models. It can also produce qualitatively comparable images when plugged into Stable Diffusion \cite{Rombach2021HighResolutionIS} for text-guided synthesis.
\end{enumerate}

%% file: 2background.tex
\section{Related work}
\label{sec:background}
\begin{figure}[t]
\centering
\includegraphics[width=\linewidth]{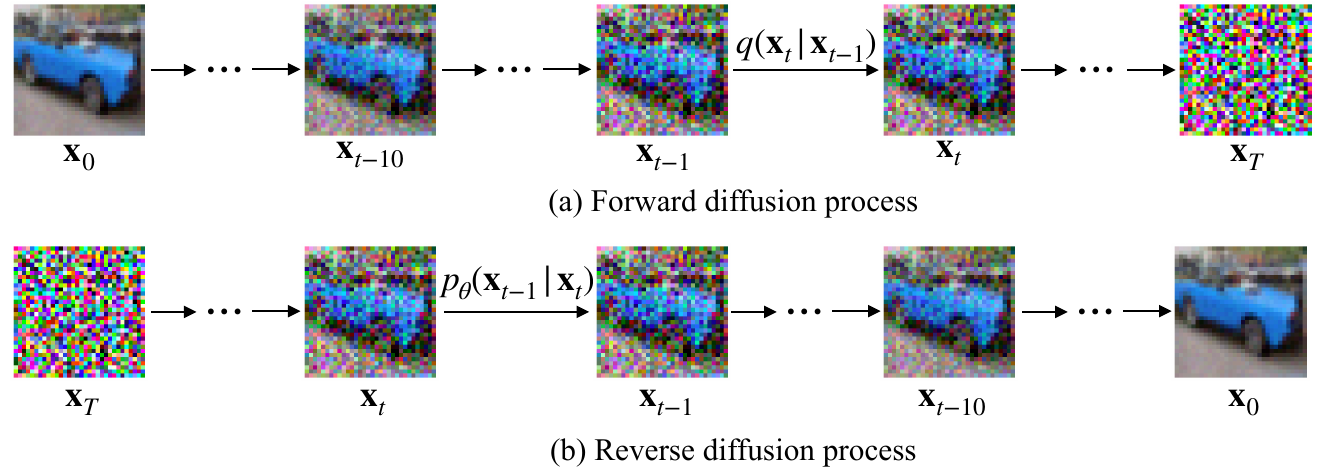}
\caption{The forward diffusion process \textbf{(a)} repeatedly adds Gaussian noise. The reverse diffusion process \textbf{(b)} uses a trained network to denoise from a standard Gaussian noise image in order to generate an image.} 
\vspace{-10pt}
\label{fig:diffusion}
\end{figure}

\noindent\textbf{Diffusion Models.}
Diffusion models generate images through a Markov chain, as illustrated in \fig{fig:diffusion}. 
A forward diffusion process adds Gaussian noise to data $\vec{x}_0 \sim q(\vec{x})$ for $T$ times, resulting in noisy samples $\vec{x}_1, ..., \vec{x}_T$:
\begin{align}
q(\vec{x}_t|\vec{x}_{t-1}) = \mathcal{N}(\vec{x}_{t}; \sqrt{1-\beta_t}\vec{x}_{t-1}, \beta_t\mathbf{I})
\end{align}
where $\beta_t \in (0,1)$ is the variance schedule that controls the strength of the Gaussian noise in each step.
When $T \rightarrow \infty$, $\vec{x}_T$ approaches an isotropic Gaussian distribution.

The reverse process removes noise from a sample from the Gaussian noise input $\vec{x}_T \sim \mathcal{N}(\vec{0}, \mathbf{I})$ to gradually generate high-fidelity images. 
However, since the real reverse conditional distribution $q(\vec{x}_{t-1}|\vec{x}_t)$ is unavailable, diffusion models sample from a learned conditional distribution: 
\begin{align} \label{eq:diff_rev}
p_\theta(\vec{x}_{t-1} | \vec{x}_t) &= \mathcal{N}(\vec{x}_{t-1}; \tilde{\vec{\mu}}_{\theta,t}(\vec{x}_t), \tilde{\beta}_t\mathbf{I}).
\end{align}
With the reparameterization trick in \cite{ho2020denoising}, the mean $\tilde{\vec{\mu}}_{\theta,t}(\vec{x}_t)$ and variance $\tilde \beta_t$ could be derived as follows:
\begin{align}
\tilde{\vec{\mu}}_{\theta,t}(\vec{x}_t) &= \frac{1}{\sqrt{\alpha_t}} (\mathbf{x}_t - \frac{1 - \alpha_t}{\sqrt{1 - \bar{\alpha}_t}} \vec{\epsilon}_{\theta,t}) \\
\tilde \beta_t &= \frac{1 - \bar{\alpha}_{t-1}}{1 - \bar{\alpha}_t} \cdot \beta_t
\end{align}
where $\alpha_t = 1 - \beta_t$, $\bar \alpha_t = \prod_{i=1}^t \alpha_i$.
We refer readers to \cite{luo2022understanding} for a more detailed introduction.

In practice, the noise at each time step $t$ are computed from $\vec{x}_t$ by a noise estimation model, with the same weights for all time steps. The UNet~\cite{ronneberger2015u} dominates the design of the noise estimation model in diffusion models~\cite{song2020denoising,Rombach2021HighResolutionIS,ramesh2022hierarchical,saharia2022photorealistic}, with some recent exploration on Transformer~\cite{Peebles2022DiT}. This work designs the PTQ method for the acceleration of the noise estimation model, especially for the common UNet.

\noindent\textbf{Accelerated diffusion process.}
Related methods include simulating the diffusion process in fewer steps by generalizing it to a non-Markovian process \cite{song2020denoising}, adjusting the variance schedule \cite{nichol2021improved}, and the use of high-order solvers to approximate diffusion generation \cite{liu2022pseudo, Bao2022AnalyticDPMAA, Lu2022DPMSolverAF, Lu2022DPMSolverFS}. Others have employed the technique of caching and reusing feature maps \cite{li2022efficient}. Efforts to distill the diffusion model into fewer time steps have also been undertaken~\cite{Salimans2022ProgressiveDF, meng2022distillation}, which have achieved notable success but involve an extremely expensive retraining process. Our work focuses on accelerating the noise estimation model inference in each step, with a training-free PTQ process.

\noindent\textbf{Post-training Quantization.}
Post-training quantization (PTQ) compresses deep neural networks by rounding elements $w$ to a discrete set of values, where the quantization and de-quantization can be formulated as: 
\begin{equation} \label{eq:1}
\hat{w} = \mathrm{s} \cdot \mathrm{clip}(\mathrm{round}(\frac{w}{s}), c_\text{min}, c_\text{max}),
\end{equation}
where $s$ denotes the quantization scale parameters, $c_\text{min}$ and $c_\text{max}$ are the lower and upper bounds for the clipping function $\mathrm{clip}(\cdot)$. These parameters can be calibrated with the weight and activation distribution estimated in the PTQ process. The operator $\mathrm{round}(\cdot)$ represents rounding, which can be either rounding-to-nearest \cite{Wu2020EasyQuantPQ,Cai2020ZeroQAN} or adaptive rounding \cite{li2021brecq}.

Previous PTQ research in classification and detection tasks focused on the calibration objective and the acquisition of calibration data. For example, EasyQuant \cite{Wu2020EasyQuantPQ} determines appropriate $c_\text{min}$ and $c_\text{max}$ based on training data, and BRECQ \cite{li2021brecq} introduces Fisher information into the objective. ZeroQ \cite{Cai2020ZeroQAN} employs a distillation technique to generate proxy input images for PTQ, and SQuant \cite{Guo2022SQuantOD} uses random samples with objectives based on sensitivity determined through the Hessian spectrum. 
For diffusion model quantization, a training dataset is not needed as the calibration data can be constructed by sampling the full-precision model with random inputs. However, the multi-time step inference of the noise estimation model brings new challenges in modeling the activation distribution. In parallel to our work, PTQ4DM~\cite{shang2022posttraining} introduces the method of Normally Distributed Time-step Calibration, generating calibration data across all time steps with a specific distribution. Nevertheless, their explorations remain confined to lower resolutions, 8-bit precision, floating-point attention activation-to-activation matmuls, and with limited ablation study on other calibration schemes. This results in worse applicability of their method to lower precisions (see Appendix~\ref{sec:ptq4dm}). Our work delves into the implications of calibration dataset creation in a holistic manner, establishing an efficient calibration objective for diffusion models. We fully quantize act-to-act matmuls, validated by experiments involving both pixel-space and latent-space diffusion models on large-scale datasets up to resolutions of $512 \times 512$.

%% file: 3method.tex
\section{Method}
\label{sec:method}
We present our method for post-training quantization of diffusion models in this section.
Different from conventionally studied deep learning models and tasks such as CNNs and VITs for classification and detection, diffusion models are trained and evaluated in a distinctive multi-step manner with a unique UNet architecture. This presents notable challenges to the PTQ process.
We analyze the challenges brought by the multi-step inference process and the UNet architecture in Section~\ref{ssec:time} and~\ref{ssec:split} respectively and describe the full Q-Diffusion PTQ pipeline in Section~\ref{ssec:PTQ}.

\subsection{Challenges under the Multi-step Denoising}
\label{ssec:time}
We identify two major challenges in quantizing models that employ multi-step inference process. Namely, we investigate the accumulation of quantization error across time steps and the difficulty of sampling a small calibration dataset to reduce the quantization error at each time step.
\vspace{-0.3cm}
\paragraph{Challenge 1: Quantization errors accumulate across time steps.}
Performing quantization on a neural network model introduces noise on the weight and activation of the well-trained model, leading to quantization errors in each layer's output. Previous research has identified that quantization errors are likely to accumulate across layers \cite{dettmers2022llm}, making deeper neural networks harder to quantize. 
In the case of diffusion models, at any time step $t$, the input of the denoising model (denoted as $\vec{x}_t$) is derived by $\vec{x}_{t+1}$, the output of the model at the previous time step $t+1$ (as depicted by \eqn{eq:diff_rev}). This process effectively multiplies the number of layers involved in the computation by the number of denoising steps for the input $\vec{x}_{t}$ at time step $t$, leading to an accumulation of quantization errors towards later steps in the denoising process.

\begin{figure}[t]
\centering
\includegraphics[width=0.9\linewidth]{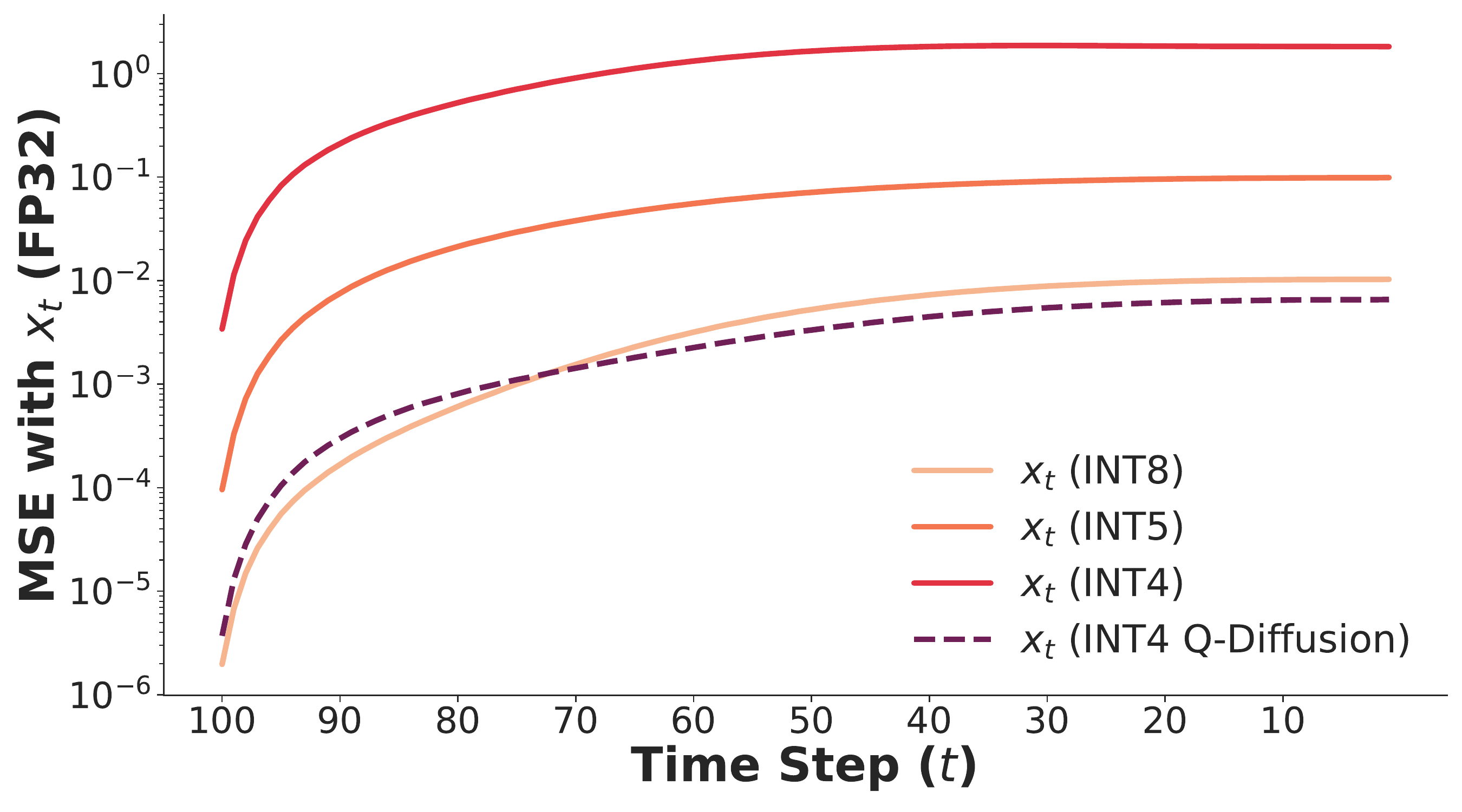}
\caption{MSE between FP32 outputs and weight-quantized outputs of different precisions with Linear Quantization and our approach across time steps. Here the data is obtained by passing a batch with 64 samples through a model trained on CIFAR-10 \cite{krizhevsky2009learning} with DDIM sampling steps 100.}
\label{tstep_mse}
\vspace{-0.3cm}
\end{figure}

\begin{figure}[!t]
\centering
\includegraphics[width=0.9\linewidth]{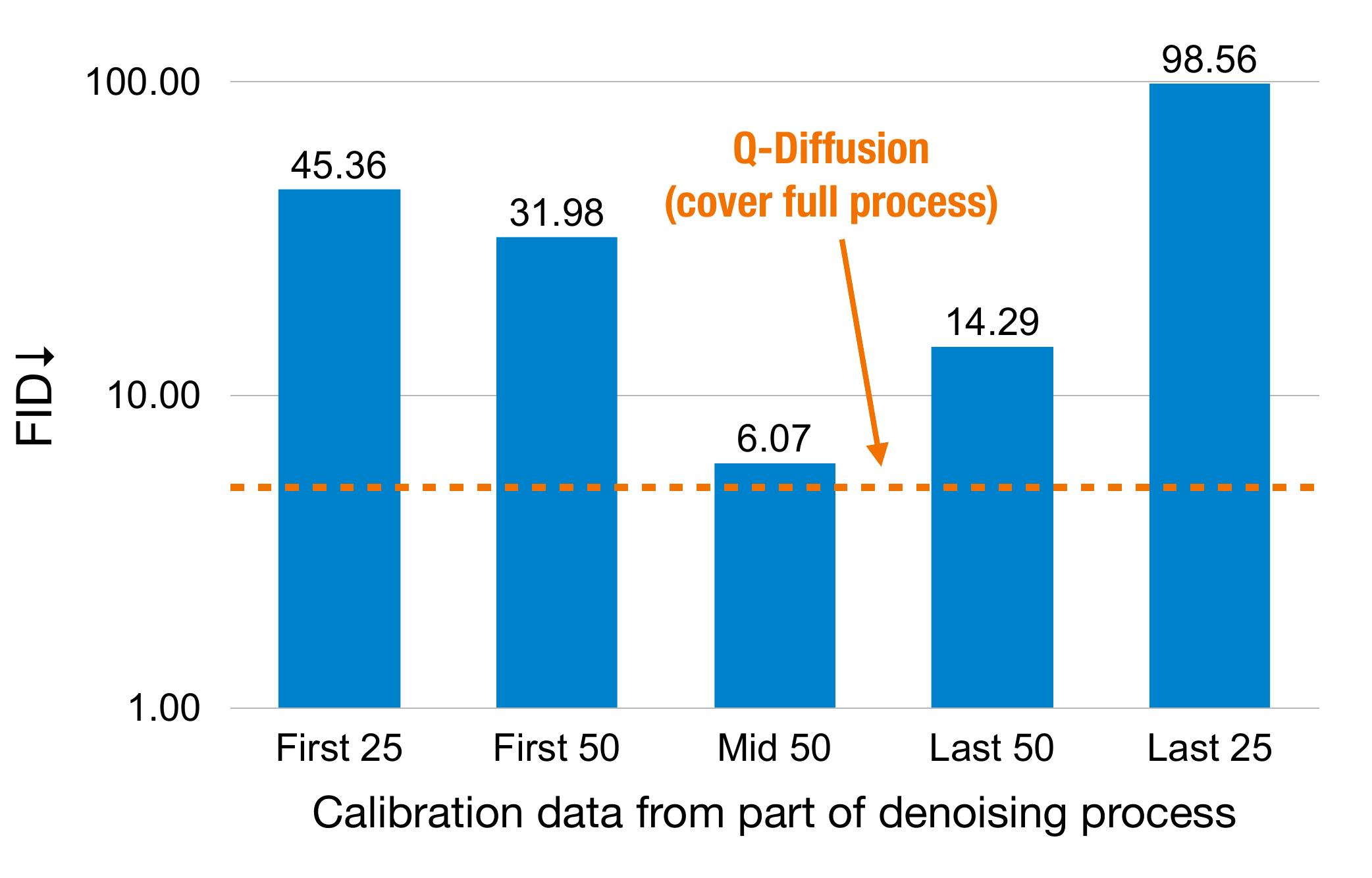}
\caption{Effects of time steps in calibration dataset creation on 4-bit weights quantization results with DDIM on \cifar. First $n$, Mid $n$, Last $n$ denotes that 5120 samples are selected uniformly from the first, middle, last $n$ time steps.}
\label{full_dist}
\vspace{-0.5cm}
\end{figure}

\begin{figure*}[t]
\centering
\vspace{-5pt}
\includegraphics[width=0.85\linewidth]{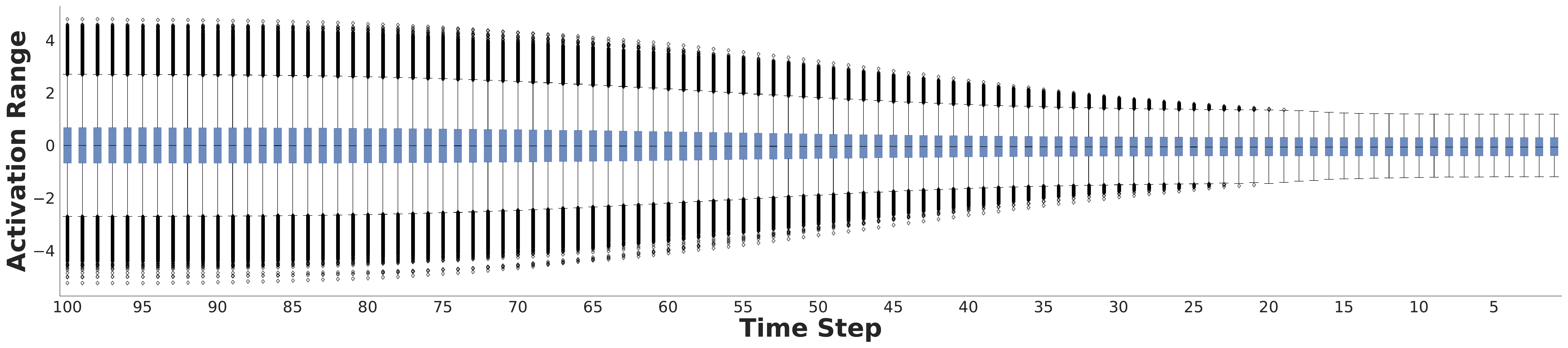}
\caption{Activation ranges of $x_t$ across all 100 time steps of FP32 DDIM model on \cifar.}
\label{act_tstep}
\end{figure*}

We run the denoising process of DDIM \cite{song2020denoising} on CIFAR-10 \cite{krizhevsky2009learning} with a sampling batch size of 64, and compare the MSE differences between the full-precision model and the model quantized to INT8, INT5, and INT4 at each time step. As shown in \fig{tstep_mse}, there is a dramatic increase in the quantization errors when the model is quantized to 4-bit, and the errors accumulate quickly through iterative denoising. This brings difficulty in preserving the performance after quantizing the model down to low precision, which requires the reduction of quantization errors at all time steps as much as possible.
\vspace{-0.3cm}
\paragraph{Challenge 2: Activation distributions vary across time steps.} 
To reduce the quantization errors at each time step, previous PTQ research \cite{Nagel2020UpOD, li2021brecq} calibrates the clipping range and scaling factors of the quantized model with a small set of calibration data. The calibration data should be sampled to resemble the true input distribution so that the activation distribution of the model can be estimated correctly for proper calibration. 
Given that the Diffusion model uses the same noise estimation network to take inputs from all time steps, determining the data sampling policy across different time steps becomes an outstanding challenge.
Here we start by analyzing the output activation distribution of the UNet model across different time steps. We conduct the same CIFAR-10 experiment using DDIM with 100 denoising steps, and draw the activations ranges of 1000 random samples among all time steps. As \fig{act_tstep} shows, the activation distributions gradually change, with neighboring time steps being similar and distant ones being distinctive.
This is also echoed by the visualized $\Vec{x}_t$ in \fig{fig:diffusion}.

The fact that the output activations distribution varies across time steps further brings challenges to quantization. Calibrating the noise estimation model using only a few time steps that do not reflect the full range of activations seen among all time steps by the noise estimation model during the denoising process can cause overfitting to the activation distribution described by those specific time steps, while not generalizing to other time steps, which hurts the overall performance.
For instance, here we try to calibrate the quantized DDIM on the CIFAR-10 dataset with data sampled from different parts of the denoising process. 
As shown in \fig{full_dist}, if we simply take 5120 samples from time steps that fall into a certain stage of the denoising process, significant performance drops will be induced under 4-bit weights quantization. Note that the case with samples taken from the middle 50 time steps caused smaller drops compared to cases with samples taken from either the first or the last $n$ time steps, and with $n$ increases, the drops are also alleviated. These results illustrate the gradual ``denoising" process as depicted in \fig{act_tstep}: the activations distribution changes gradually throughout time steps, with the middle part capturing the full range to some degree, while parts of the distant endpoints differing the most.
To recover the performance of the quantized diffusion models, we need to select calibration data in a way that comprehensively takes into account the distributions of the output of different time steps.

\subsection{Challenges on Noise Estimation Model Quantization}
\label{ssec:split}

\begin{figure*}[h]
\centering
\vspace{-5pt}
\includegraphics[width=0.8\linewidth]{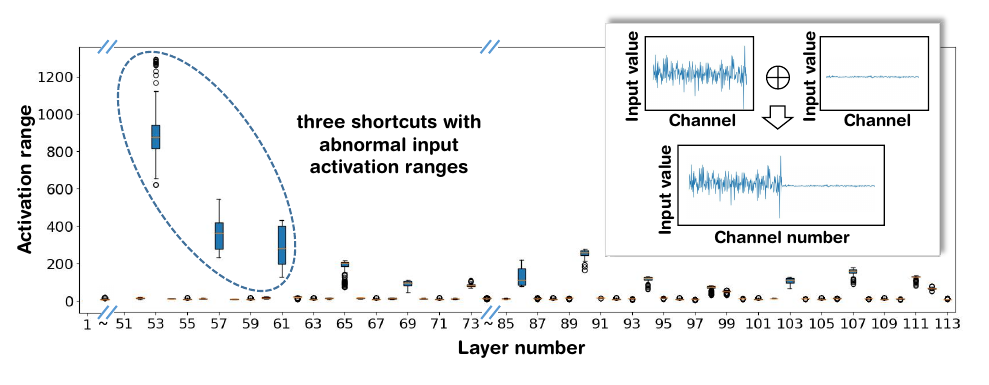}
\vspace{-20pt}
\caption{Activation ranges of DDIM's FP32 outputs across layers averaging among all time steps. We point out three shortcuts with the largest input activation ranges compared to other neighboring layers. Figures in the dashed box illustrate concatenation along channels. $\oplus$ denotes the concatenation operation.}
\label{ddim_act_layers}
\end{figure*}

Most diffusion models (Imagen~\cite{saharia2022photorealistic}, Stable Diffusion~\cite{Rombach2021HighResolutionIS}, VDMs~\cite{ho2022video}) adopt UNets as denoising backbones that downsample and upsample latent features. Although recent studies show that transformer architectures are also capable of serving as the noise estimation backbone~\cite{Peebles2022DiT}, 
convolutional UNets are still the de facto choice of architecture today.
UNets utilize shortcut layers to merge concatenated deep and shallow features and transmit them to subsequent layers.  Through our analysis presented in \fig{ddim_act_layers}, we observe that input activations in shortcut layers exhibit abnormal value ranges in comparison to other layers. Notably, the input activations in DDIM's shortcut layers can be up to 200 times larger than other neighboring layers. 

\begin{figure*}[h]
\centering
\vspace{-5pt}
\includegraphics[width=1.0\linewidth]{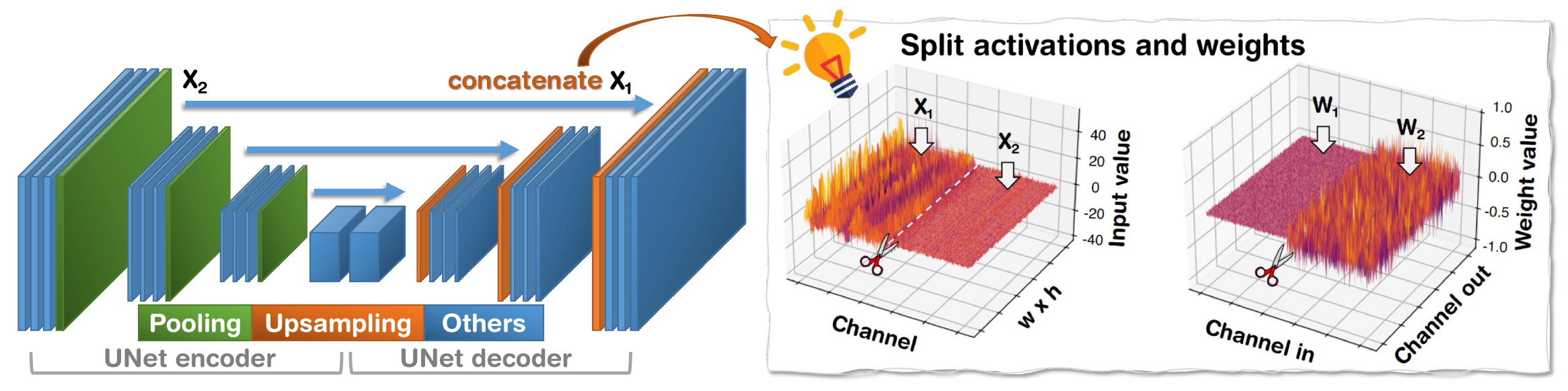}
\caption{(Left) The typical UNet architecture with shortcut layers that concatenate features from the deep and shallow layers. (Right) The ranges of activations from the deep ($X_1$) and shallow ($X_2$) feature channels vary significantly, which also results in a bimodal weight distribution in the corresponding channels.}
\label{split_unet}
\vspace{-5pt}
\end{figure*}

To analyze the reason for this, we visualize the weight and activation tensor of a DDIM shortcut layer. As demonstrated in the dashed box in \fig{ddim_act_layers}, the ranges of activations from the deep feature channels ($X_1$) and shallow feature channels ($X_2$) being concatenated together vary significantly, which also resulted in a bimodal weight distribution in the corresponding channels (see also \fig{split_unet}).
Naively quantizing the entire weight and activation distribution with the same quantizer will inevitably lead to large quantization errors.

\subsection{Post-Training Quantization of Diffusion model}
\label{ssec:PTQ}

 We propose two techniques: \textit{time step-aware calibration data sampling} and \textit{shortcut-splitting quantization} to tackle the challenges identified in the previous sections respectively.
 
 \subsubsection{Time step-aware calibration} Since the output distributions of consecutive time steps are often very similar, we propose to randomly sample intermediate inputs uniformly in a fixed interval across all time steps to generate a small calibration set. This effectively balances the size of the calibration set and its representation ability of the distribution across all time steps. 
 Empirically, we have found that the sampled calibration data can recover most of the INT4 quantized models' performance after the calibration, making it an effective sampling scheme for calibration data collection for quantization error correction. 

 To calibrate the quantized model, we divide the model into several reconstruction blocks \cite{li2021brecq}, and iteratively reconstruct outputs and tune the clipping range and scaling factors of weight quantizers in each block with adaptive rounding \cite{Nagel2020UpOD} to minimize the mean squared errors between the quantized and full precision outputs. We define a core component that contains residual connections in the diffusion model UNet as a block, such as a Residual Bottleneck Block or a Transformer Block. Other parts of the model that do not satisfy this condition are calibrated in a per-layer manner. This technique has been shown to improve the performance compared to fully layer-by-layer calibration since it address the inter-layer dependencies and generalization better \cite{li2021brecq}. For activation quantization, since activations are constantly changing during inference, doing adaptive rounding is infeasible. Therefore, we only adjust the step sizes of activation quantizers according to to~\cite{Esser2020LEARNED}. The overall calibration workflow is described in Alg. \ref{alg:calib}.
\begin{algorithm}
\caption{Q-Diffusion Calibration} \label{alg:qdiff}
\textbf{Require:} Pretrained full precision diffusion model and the quantized diffusion model [$W_\theta$, $\hat{W}_\theta$]  \\
\textbf{Require:} Empty calibration dataset $\mathcal{D}$ \\
\textbf{Require:} Number of denoising sampling steps $T$ \\
\textbf{Require:} Calibration sampling interval $c$, amount of calibration data per sampling step $n$ 
\begin{algorithmic}
\FOR {$t = 1, \ldots, T$ time step}
    \IF{t \% c = 0}
        \STATE Sample $n$ intermediate inputs $\vec{x}^{(1)}_t, \ldots, \vec{x}^{(n)}_t$ randomly at $t$ from $W_\theta$ and add them to $\mathcal{D}$
    \ENDIF
\ENDFOR
\FOR {all $i = 1, \ldots, N$ blocks}
    \STATE Update the weight quantizers of the $i$-th block in $\hat{W}_\theta$ with $\mathcal{D}$ and $W_\theta$
\ENDFOR
\IF{do activation quantization}
    \FOR {all $i = 1, \ldots, N$ blocks}
        \STATE Update the activation quantizers step sizes of the $i$-th block with $\hat{W}_\theta$, $W_\theta$, $\mathcal{D}$.
    \ENDFOR
\ENDIF
\end{algorithmic}
\label{alg:calib}
\end{algorithm}

\subsubsection{Shortcut-splitting quantization} To address the abnormal activation and weight distributions in shortcut layers, we propose a ``split" quantization technique that performs quantization prior to concatenation, requiring negligible additional memory or computational resources. This strategy can be employed for both activation and weight quantization in shortcut layers, and is expressed mathematically as follows:
\begin{align} 
\label{eq:split}
\mathcal{Q}_X(X) &= \mathcal{Q}_{X_1}(X_1) \oplus \mathcal{Q}_{X_2}(X_2) \\
\mathcal{Q}_W(W) &= \mathcal{Q}_{W_1}(W_1) \oplus \mathcal{Q}_{W_2}(W_2)
\end{align}
\begin{equation}
\begin{split}
\mathcal{Q}_{X}(X)\mathcal{Q}_{W}(W) = &\mathcal{Q}_{X_1}(X_{1})\mathcal{Q}_{W_1}(W_{1}) \\ &+ \mathcal{Q}_{X_2}(X_{2})\mathcal{Q}_{W_2}(W_{2})    
\end{split}
\end{equation}
where $X\in \mathbb{R}^{w\times h\times c_{in}}$ and $W\in \mathbb{R}^{c_{in}\times c_{out}}$ are the input activation and layer weight, which can be naturally split into $X_{1}\in \mathbb{R}^{w\times h\times c_{1}}$, $X_{2}\in \mathbb{R}^{w\times h\times c_{2}}$, $W_{1}\in \mathbb{R}^{c_{1}\times c_{out}}$, and $W_{2}\in \mathbb{R}^{c_{2}\times c_{out}}$, respectively. $c_{1}$ and $c_{2}$ are determined by the concatenation operation. $\mathcal{Q}(\cdot)$ denotes the quantization operator and $\oplus$ denotes the concatenation operator.


%% file: 4experiment.tex
\section{Experiments}
\label{sec:exp}
\subsection{Experiments Setup}
In this section, we evaluate the proposed Q-Diffusion framework on pixel-space diffusion model DDPM \cite{ho2020denoising} and latent-space diffusion model Latent Diffusion \cite{Rombach2021HighResolutionIS} for unconditional image generation. We also visualize the images generated by Q-Diffusion on Stable Diffusion. To the best of our knowledge, there is currently no published work done on diffusion model quantization. Therefore, we report the basic channel-wise Linear Quantization (i.e., \eqn{eq:1}) as a baseline. We also re-implement the state-of-the-art data-free PTQ method SQuant~\cite{Guo2022SQuantOD} and include the results for comparison. Furthermore, we apply our approach to text-guided image synthesis with Stable Diffusion~\cite{Rombach2021HighResolutionIS}. Experiments show that our approach can achieve competitive generation quality to the full-precision scenario on all tasks, even under INT4 quantization for weights. 

\begin{figure*}[!ht]
  \begin{center}
  \begin{minipage}[c]{0.24\linewidth}
    \centering
    \footnotesize Bedroom Q-Diffusion \\ (W4A8) \\
    \vspace{0.1cm}
    \includegraphics[width=0.99\textwidth]{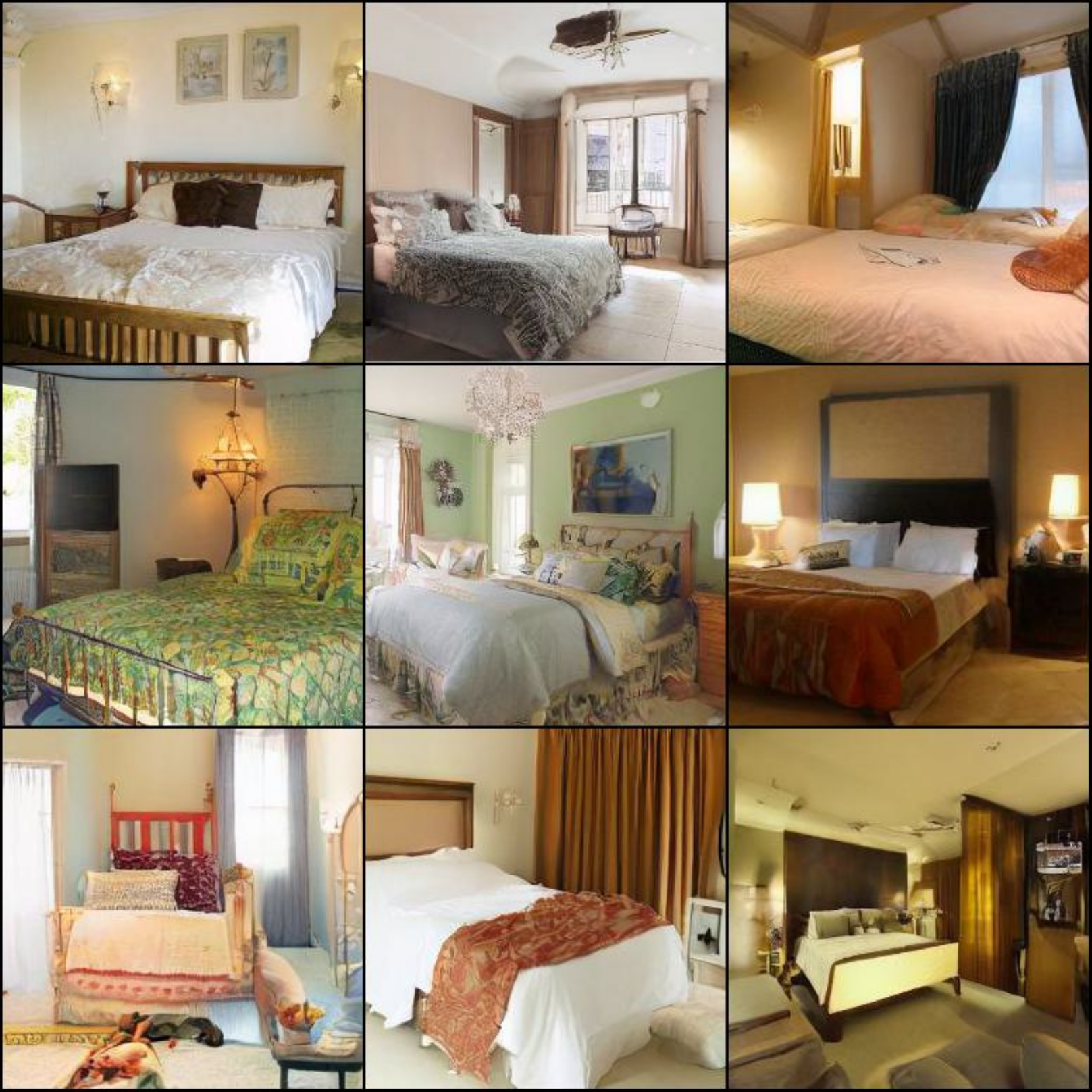}
    \label{fig:bedroom_qdiff}
  \end{minipage}\hfill
  \begin{minipage}[c]{0.24\linewidth}
    \centering
    \footnotesize Bedroom Linear Quantization \\ (W4A8) \\
    \vspace{0.1cm}
    \includegraphics[width=0.99\textwidth]{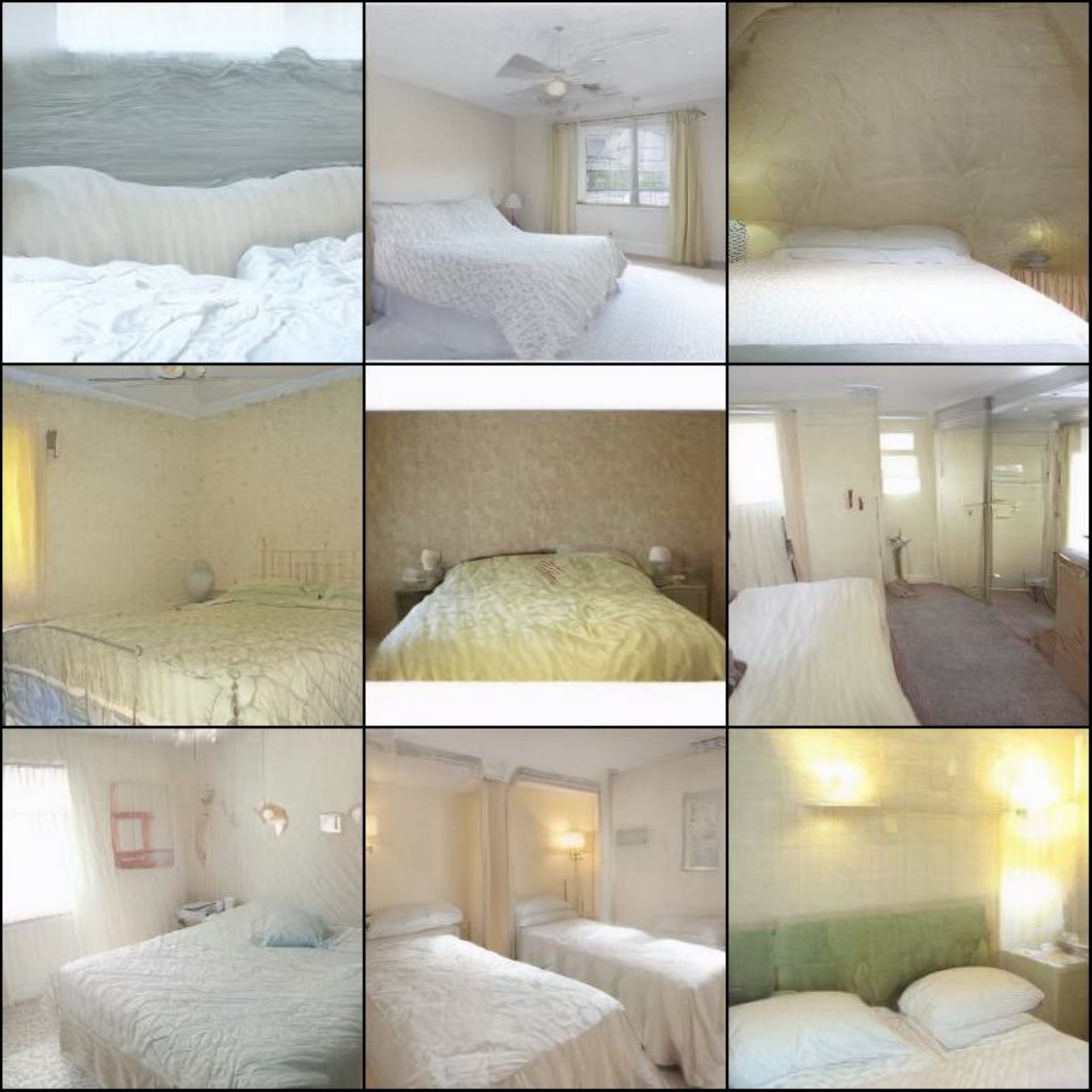}
    \label{fig:bedroom_lq}
  \end{minipage}\hfill
  \begin{minipage}[c]{0.24\linewidth}
    \centering
    \footnotesize Church Q-Diffusion \\ (W4A8) \\
    \vspace{0.1cm}
    \includegraphics[width=0.99\textwidth]{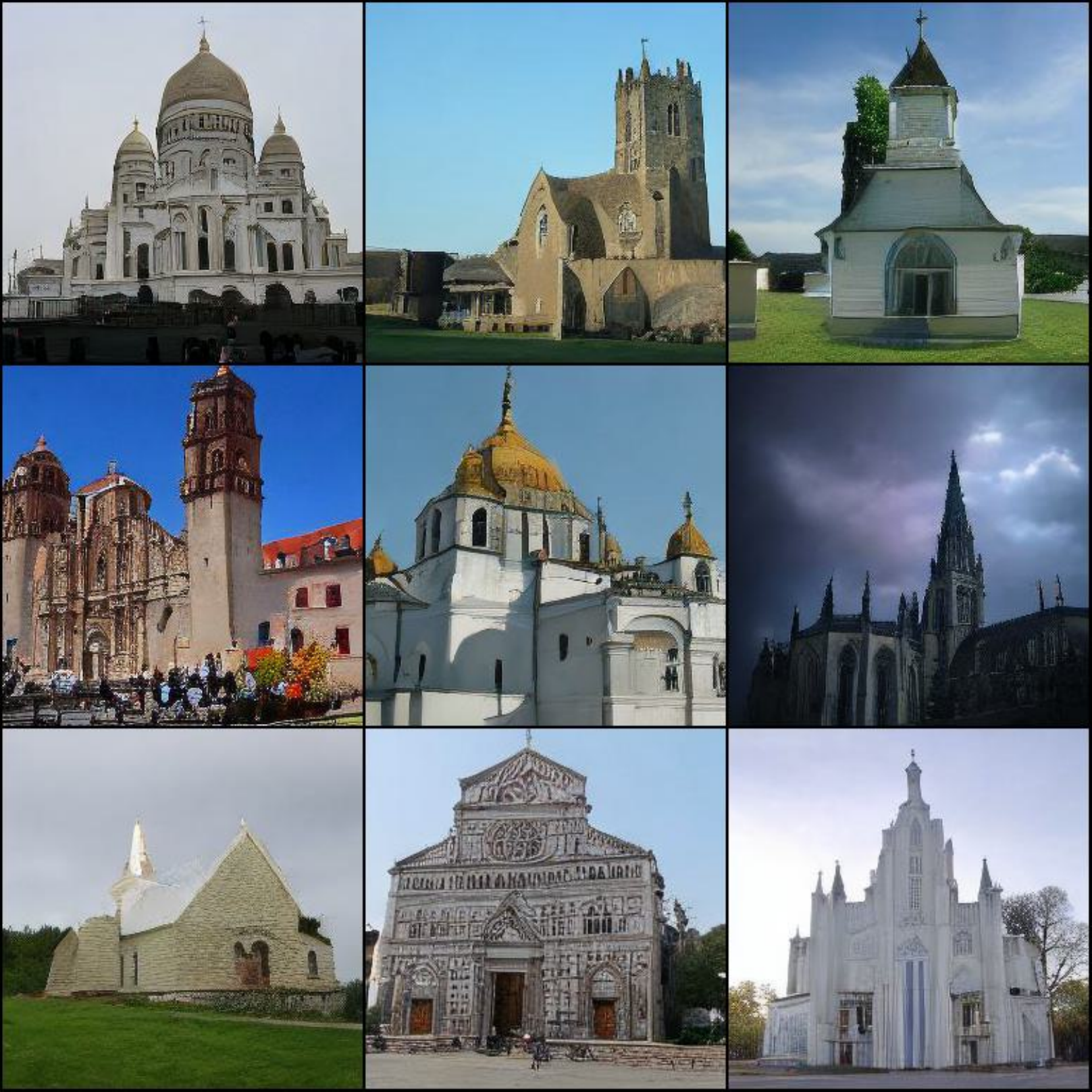}
    \label{fig:church_qdiff}
  \end{minipage}\hfill
  \begin{minipage}[c]{0.24\linewidth}
    \centering
    \footnotesize Church Linear Quantization \\ (W4A8) \\
    \vspace{0.1cm}
    \includegraphics[width=0.99\textwidth]{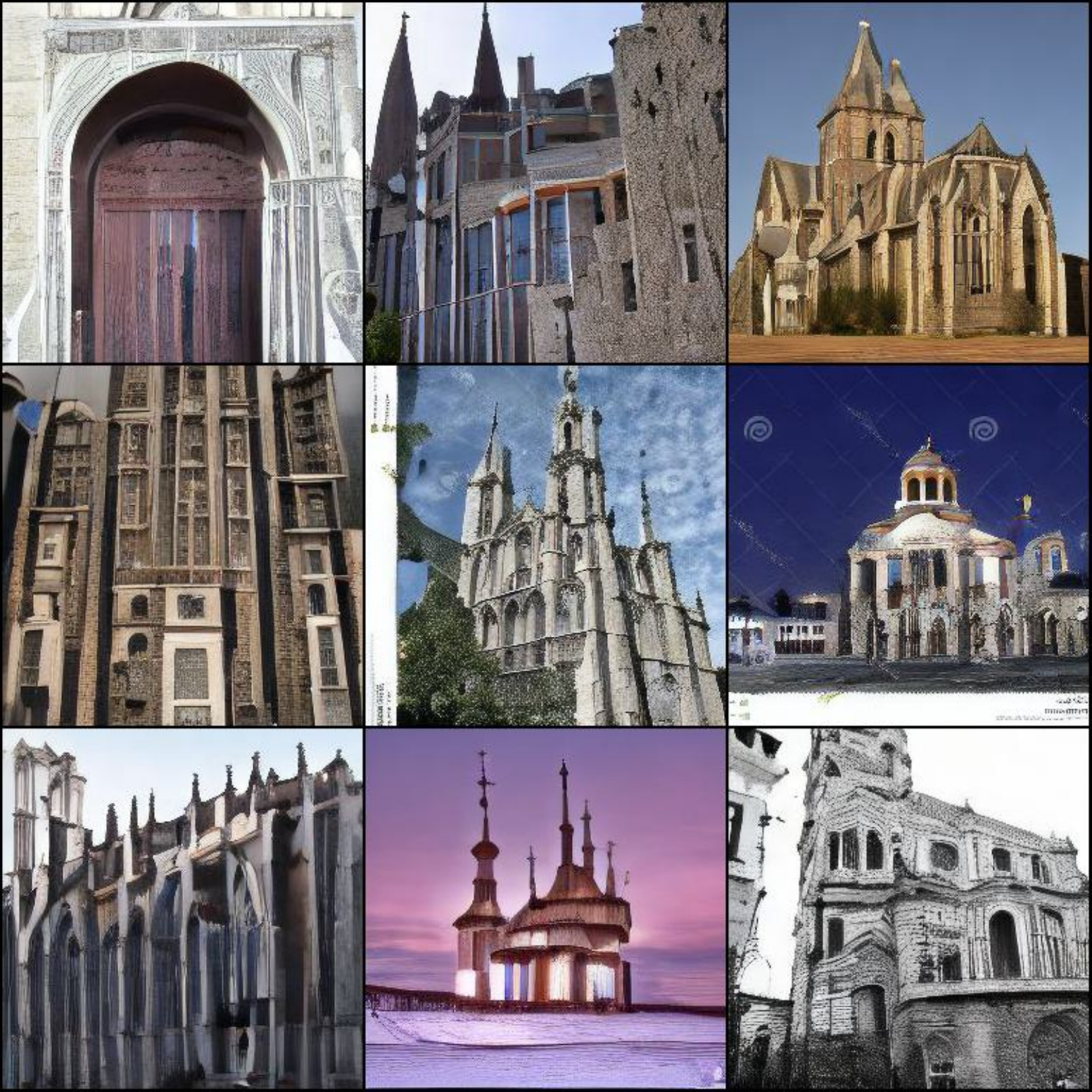}
    \label{fig:church_lq}
  \end{minipage}
  \vspace{-15pt}
  \caption{256 $\times$ 256 unconditional image generation results using Q-Diffusion and Linear Quantization under W4A8 precision.}
  \label{fig:uncond_results}
  \end{center}
  \vspace{-5pt}
\end{figure*}

\subsection{Unconditional Generation}
We conducted evaluations using the 32 $\times$ 32 CIFAR-10 \cite{krizhevsky2009learning}, $256\times256$ LSUN Bedrooms, and $256\times256$ LSUN  Church-Outdoor \cite{yu2015lsun}. We use the pretrained DDIM sampler \cite{song2020denoising} with 100 denoising time steps for CIFAR-10 experiments and Latent Diffusion (LDM) \cite{Rombach2021HighResolutionIS} for the higher resolution LSUN experiments. We evaluated the performance in terms of Frechet Inception Distance (FID) \cite{heusel2017fid} and additionally evaluated the Inception Score (IS) \cite{salimans2016is} for CIFAR-10 results, since IS is not an accurate reference for datasets that differ significantly from ImageNet's domain and categories. The results are reported in \tbl{tab:cifar_results}- \ref{tab:church_results} and \fig{fig:uncond_results}, where Bops is calculated for one denoising step without considering the decoder compute cost for latent diffusion.

The experiments show that \name significantly preserves the image generation quality and outperforms Linear Quantization by a large margin  for all resolutions and types of diffusion models tested when the number of bits is low. Although 8-bit weight quantization has almost no performance loss compared to FP32 for both Linear Quantization and our approach, the generation quality with Linear Quantization drops drastically under 4-bit weight quantization. In contrast, \name still preserves most of the perceptual quality with at most $2.34$ increase in FID and imperceptible distortions in produced samples.

\begin{table}[tb]
    \centering
    \caption{Quantization results for unconditional image generation with DDIM on CIFAR-10 (32 $\times$ 32).}
    \vspace{-5pt}
\label{tab:cifar_results}
\resizebox{0.9\linewidth}{!}
{
    \begin{tabular}{l@{\hskip 0.02in}r@{\hskip 0.1in}r@{\hskip 0.1in}r@{\hskip 0.1in}rr}
    \toprule
    Method  & Bits (W/A) & Size (Mb) & GBops & FID$\downarrow$ & IS$\uparrow$   \\
    \midrule
    Full Precision     & 32/32 & 143.2& 6597 & 4.22 &  9.12 \\
    \midrule
    Linear Quant & 8/32  & 35.8 & 2294 & 4.71 & 8.93 \\
    SQuant       & 8/32  & 35.8 & 2294 & 4.61 & 8.99  \\
    \ourcell \name & \ourcell 8/32 & \ourcell 35.8 & \ourcell 2294 &  \ourcell \textbf{4.27} & \ourcell \textbf{9.15}  \\
    \midrule
    Linear Quant & 4/32  & 17.9 & 1147 & 141.47 & 4.20  \\
    SQuant       & 4/32  & 17.9 & 1147 & 160.40 & 2.91  \\
    \ourcell \name & \ourcell 4/32  & \ourcell 17.9 & \ourcell 1147 & \ourcell \textbf{5.09}  &  \ourcell \textbf{8.78} \\
    \midrule
    Linear Quant & 8/8  & 35.8 & 798 & 118.26 &  5.23 \\
    SQuant       &  8/8 & 35.8 & 798 & 464.69 &  1.17 \\
    \ourcell \name & \ourcell 8/8 & \ourcell 35.8 & \ourcell 798 & \ourcell \textbf{3.75}  & \ourcell \textbf{9.48}  \\
    \midrule
    Linear Quant & 4/8 & 17.9 & 399 & 188.11  & 2.45  \\
    SQuant       & 4/8 & 17.9 & 399 & 456.21 &  1.16 \\
    \ourcell \name & \ourcell 4/8 & \ourcell 17.9 & \ourcell 399 & \ourcell \textbf{4.93}  & \ourcell \textbf{9.12}  \\
    \bottomrule
    \end{tabular}
    }
\vspace{-15pt}
\end{table}

\begin{table}[tb]
    \centering
    \caption{Quantization results for unconditional image generation with LDM-4 on LSUN-Bedrooms (256 $\times$ 256). The downsampling factor for the latent space is 4.}
     \vspace{-5pt}
\label{tab:bedroom_results}
\resizebox{0.9\linewidth}{!}
{
    \begin{tabular}{l@{\hskip 0.02in}rrrr}
    \toprule
    Method  & Bits (W/A) & Size (Mb) & TBops & FID$\downarrow$ \\ 
    \midrule
    Full Precision  & 32/32  & 1096.2 & 107.17 & 2.98 \\
    \midrule
    Linear Quant           & 8/32  & 274.1 & 37.28 & 3.02 \\
    SQuant                 & 8/32  & 274.1 & 37.28 & \textbf{2.94} \\
    \ourcell \name   & \ourcell 8/32 & \ourcell 274.1 & \ourcell 37.28 & \ourcell 2.97 \\
    \midrule
    Linear Quant           & 4/32  & 137.0 & 18.64 & 82.69 \\
    SQuant                 & 4/32  & 137.0 & 18.64 & 149.97 \\
    \ourcell \name   & \ourcell 4/32 & \ourcell 137.0 & \ourcell 18.64 & \ourcell \textbf{4.86} \\
    \midrule
    Linear Quant           & 8/8  & 274.1 & 12.97 & 6.69 \\
    SQuant                 & 8/8  & 274.1 & 12.97 & 4.92 \\
    \ourcell \name   & \ourcell 8/8 & \ourcell 274.1 & \ourcell 12.97 & \ourcell \textbf{4.40} \\
    \midrule
    Linear Quant           & 4/8  & 137.0 & 6.48 & 24.86 \\
    SQuant                 & 4/8  & 137.0 & 6.48 & 95.92 \\
    \ourcell \name   & \ourcell 4/8 & \ourcell 137.0 & \ourcell 6.48 & \ourcell \textbf{5.32} \\
    \bottomrule
    \end{tabular}
    }
\vspace{-5pt}
\end{table}

\begin{table}[tb]
    \centering
    \caption{Quantization results for unconditional image generation with LDM-8 on LSUN-Churches (256 $\times$ 256). The downsampling factor for the latent space is 8.}
     \vspace{-5pt}
    \label{tab:church_results}
    \resizebox{0.9\linewidth}{!}
{
    \begin{tabular}{l@{\hskip 0.019in}rrrr}
    \toprule
    Method  & Bits (W/A) & Size (Mb) & TBops & FID$\downarrow$ \\ 
    \midrule
    Full Precision  & 32/32  & 1179.9 & 22.17 & 4.06 \\ 
    \midrule
    Linear Quant           & 8/32  & 295.0 & 10.73 & \textbf{3.84}  \\
    SQuant                 & 8/32  & 295.0 & 10.73 & 4.01 \\
    \ourcell \name   & \ourcell 8/32 & \ourcell 295.0 & \ourcell 10.73 & \ourcell 4.03  \\ 
    \midrule
    Linear Quant           & 4/32  & 147.5 & 5.36 & 32.54  \\
    SQuant                 & 4/32  & 147.5 & 5.36 & 33.77 \\
    \ourcell \name   & \ourcell 4/32  & \ourcell 147.5 & \ourcell 5.36 & \ourcell \textbf{4.45} \\
    \midrule
    Linear Quant           & 8/8  & 295.0 & 2.68 & 14.62 \\
    SQuant                 &  8/8 & 295.0 & 2.68 & 54.15 \\
    \ourcell \name   & \ourcell 8/8 & \ourcell 295.0 & \ourcell 2.68 & \ourcell \textbf{3.65} \\
    \midrule
    Linear Quant           & 4/8  & 147.5 & 1.34 & 14.92  \\
    SQuant                 &  4/8 & 147.5 & 1.34 & 24.50 \\
    \ourcell \name   & \ourcell 4/8 & \ourcell 147.5 & \ourcell 1.34 & \ourcell \textbf{4.12} \\
    \bottomrule
    \end{tabular}
    }
\end{table}

\begin{figure*}[!ht]
  \begin{center}
  \vspace{-10pt}
  \begin{minipage}[c]{0.32\linewidth}
    \centering
    \footnotesize Full Precision \\
    \vspace{0.1cm}
    \includegraphics[width=0.99\textwidth]{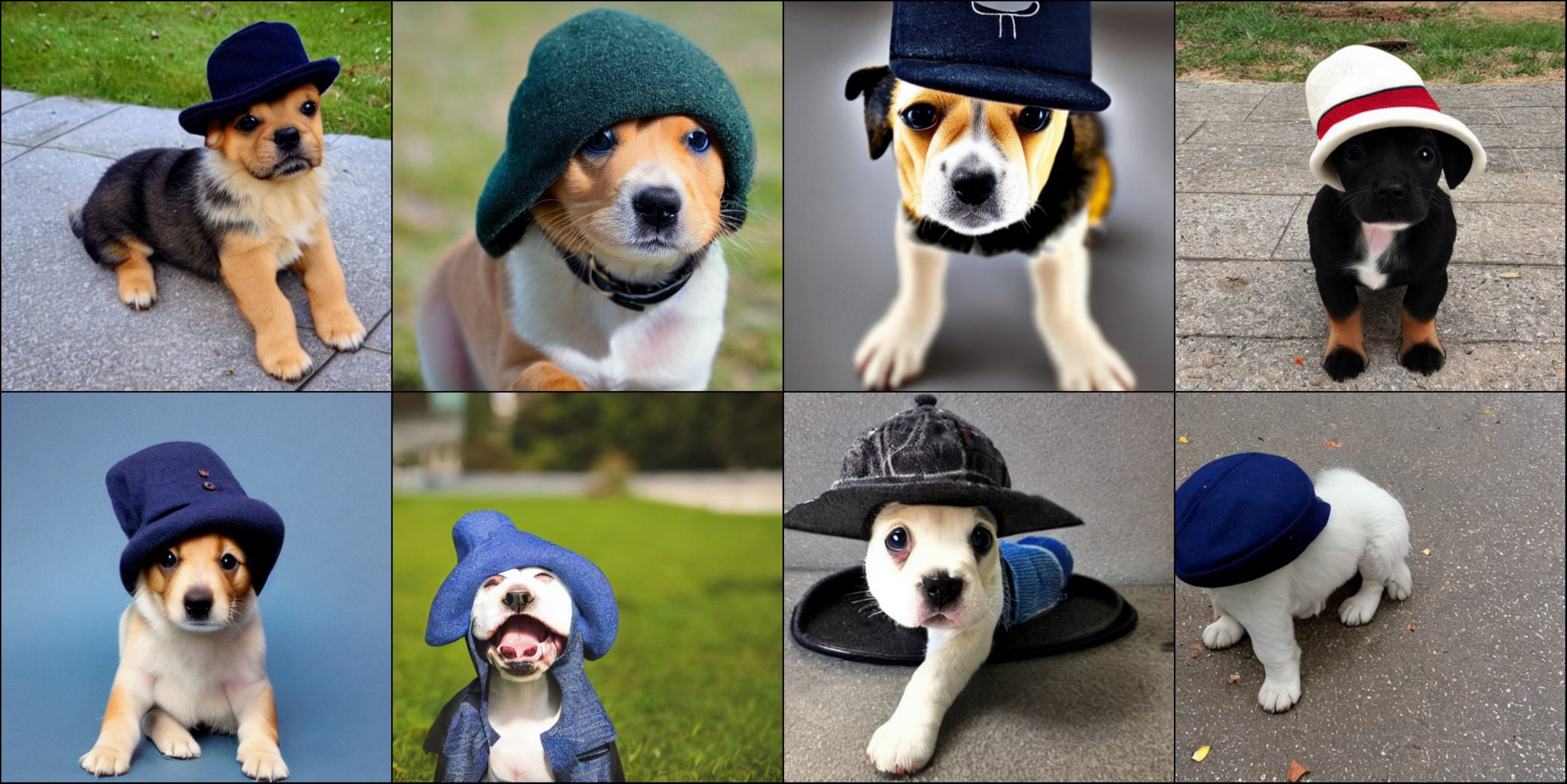}
    \label{fig:sd_base}
  \end{minipage}\hfill
  \begin{minipage}[c]{0.32\linewidth}
    \centering
    \footnotesize \name (W4A8) \\
    \vspace{0.1cm}
    \includegraphics[width=0.99\textwidth]{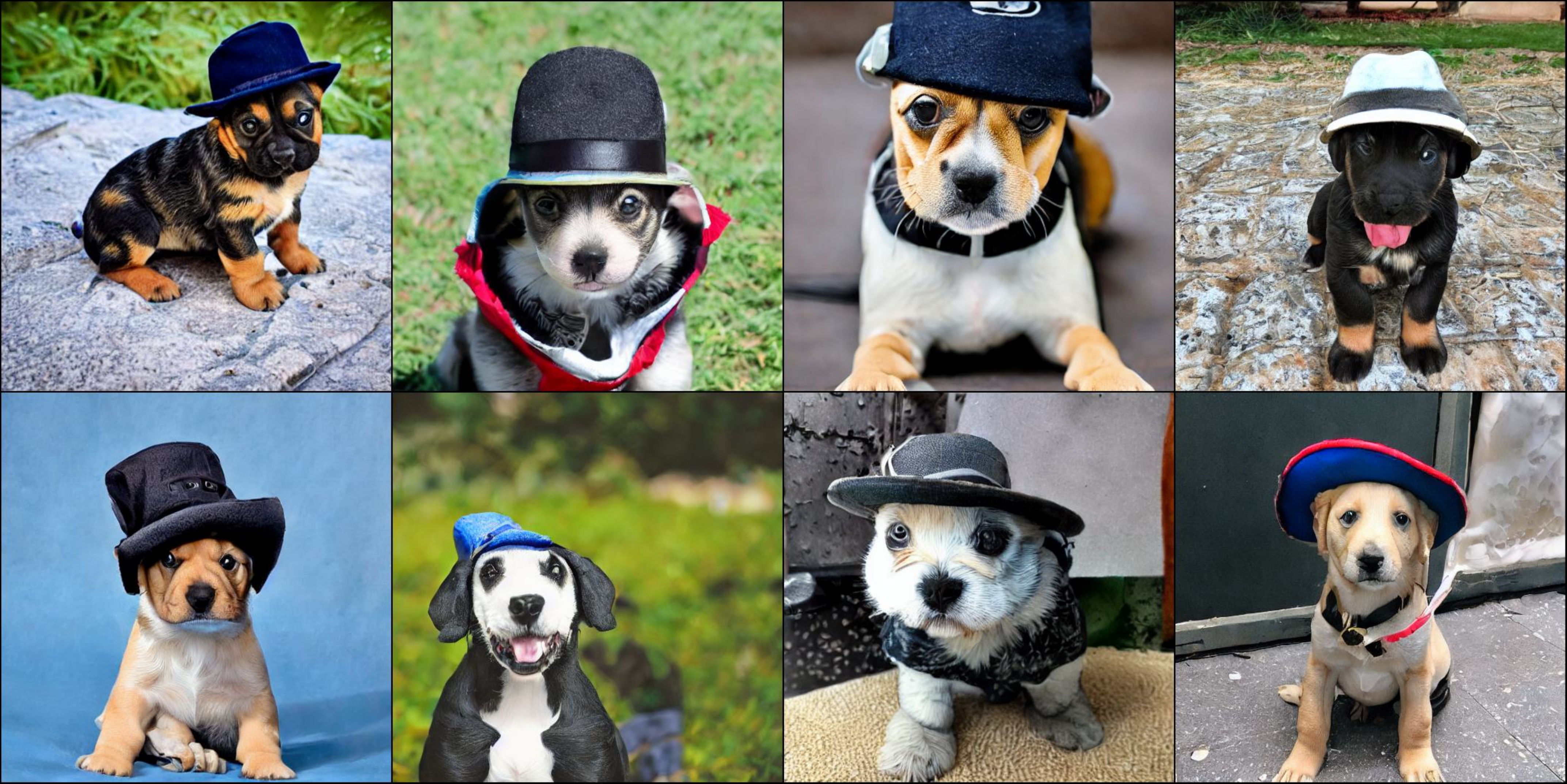}
    \label{fig:sd_qdiff}
  \end{minipage}\hfill
  \begin{minipage}[c]{0.32\linewidth}
    \centering
    \footnotesize Linear Quantization (W4A8) \\
    \vspace{0.1cm}
    \includegraphics[width=0.99\textwidth]{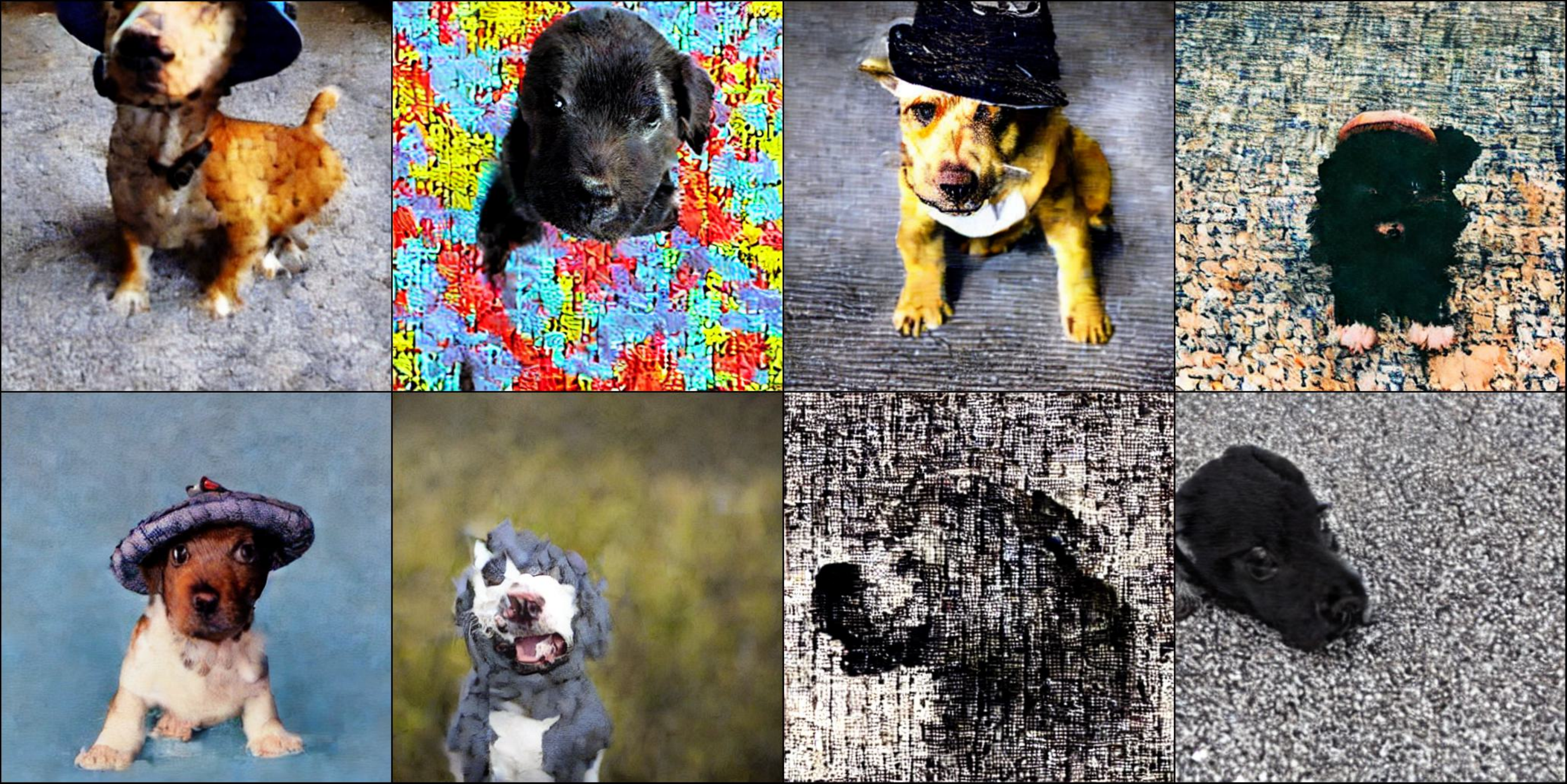}
    \label{fig:sd_lq}
  \end{minipage}
  \vspace{-0.15cm}
  \caption{Stable Diffusion 512 $\times$ 512 text-guided image synthesis results using Q-Diffusion and Linear Quantization under W4A8 precision with prompt \textit{A puppy wearing a hat}.}
  \vspace{-0.05cm}
  \label{fig:sd_results}
  \end{center}
\end{figure*}
\subsection{Text-guided Image Generation}
We evaluate Q-Diffusion on Stable Diffusion pretrained on subsets of 512 $\times$ 512 LAION-5B for text-guided image generation. Following \cite{Rombach2021HighResolutionIS}, we sample text prompts from the MS-COCO \cite{Lin2014MicrosoftCC} dataset to generate a calibration dataset with texts condition using \algo{alg:qdiff}. In this work, we fix the guidance strength to the default 7.5 in Stable Diffusion as the trade-off between sample quality and diversity. Qualitative results are shown in \fig{fig:sd_results}.
Compared to Linear Quantization, our Q-Diffusion provides higher-quality images with more realistic details and better demonstration of the semantic information. Similar performance gain is also observed in other random samples showcased in Appendix~\ref{sec:add}, and quantitatively reported in Appendix~\ref{sec:sd_score}. The output of the W4A8 Q-Diffusion model largely resembles the output of the full precision model. Interestingly, we find some diversity in the lower-level semantics between the Q-Diffusion model and the FP models, like the heading of the horse or the shape of the hat. We leave it to future work to understand how quantization contributes to the diversity.

\subsection{Ablation Study}

\paragraph{Effects of Sampling Strategies}
\begin{figure}[th]
\centering
\vspace{-5pt}
\includegraphics[width=0.9\linewidth]{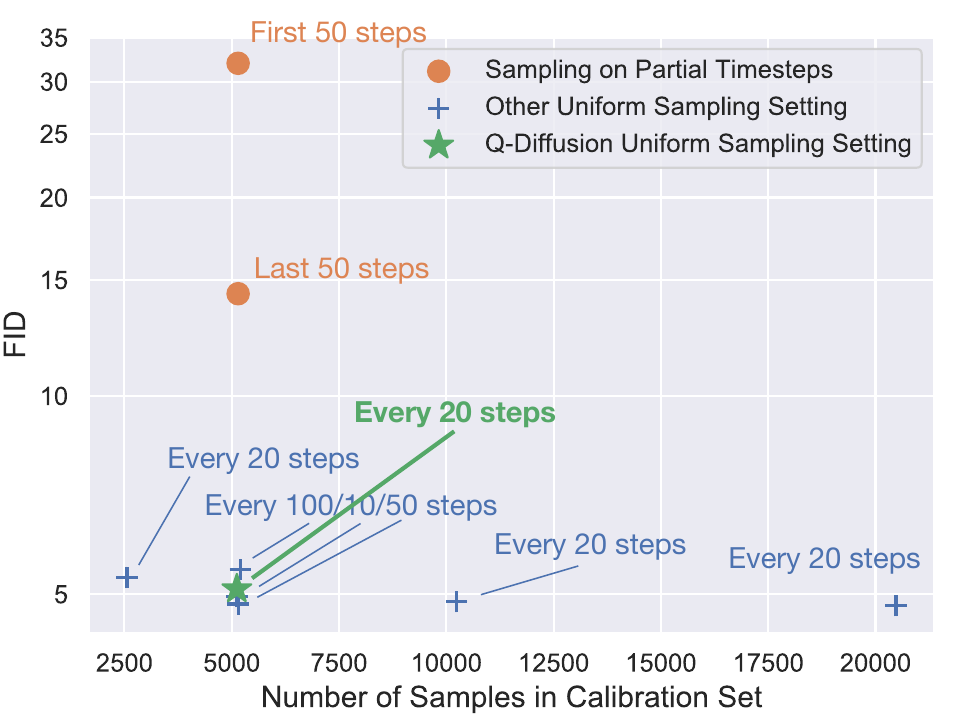}
\vspace{-10pt}
\caption{Uniform sampling strategies which cover all time steps are better than strategies that cover only a part of the time steps, as in Fig.~\ref{full_dist}. Furthermore, adjusting the sampling techniques within uniform sampling, such as tuning the sampling interval and the number of samples, has a marginal effect on the performance of the quantized model.}
\label{fig:ablation}
\vspace{-5pt}
\end{figure}

To analyze the effect of different sampling strategies for calibration in detail, we implemented multiple variants of our method using different sampling strategies. We then evaluated the quality of the models quantized by each variant. We experimented with varying numbers of time steps used for sampling and samples used for calibration. In addition to calibration sets from uniform timestep intervals, we also employed sampling at the first 50 and last 50 steps. As in Figure~\ref{fig:ablation}, uniform sampling that spans all time steps results in superior performance compared to sampling from only partial time steps. Furthermore, adjusting the sampling hyperparams, including using more calibration samples, does not significantly improve the performance. Therefore, we simply choose to sample uniformly every 20 steps for a total of 5,120 samples for calibration, resulting in a high-quality quantized model with low computational costs during quantization.

\begin{figure}[h]
\centering
\includegraphics[width=\linewidth]{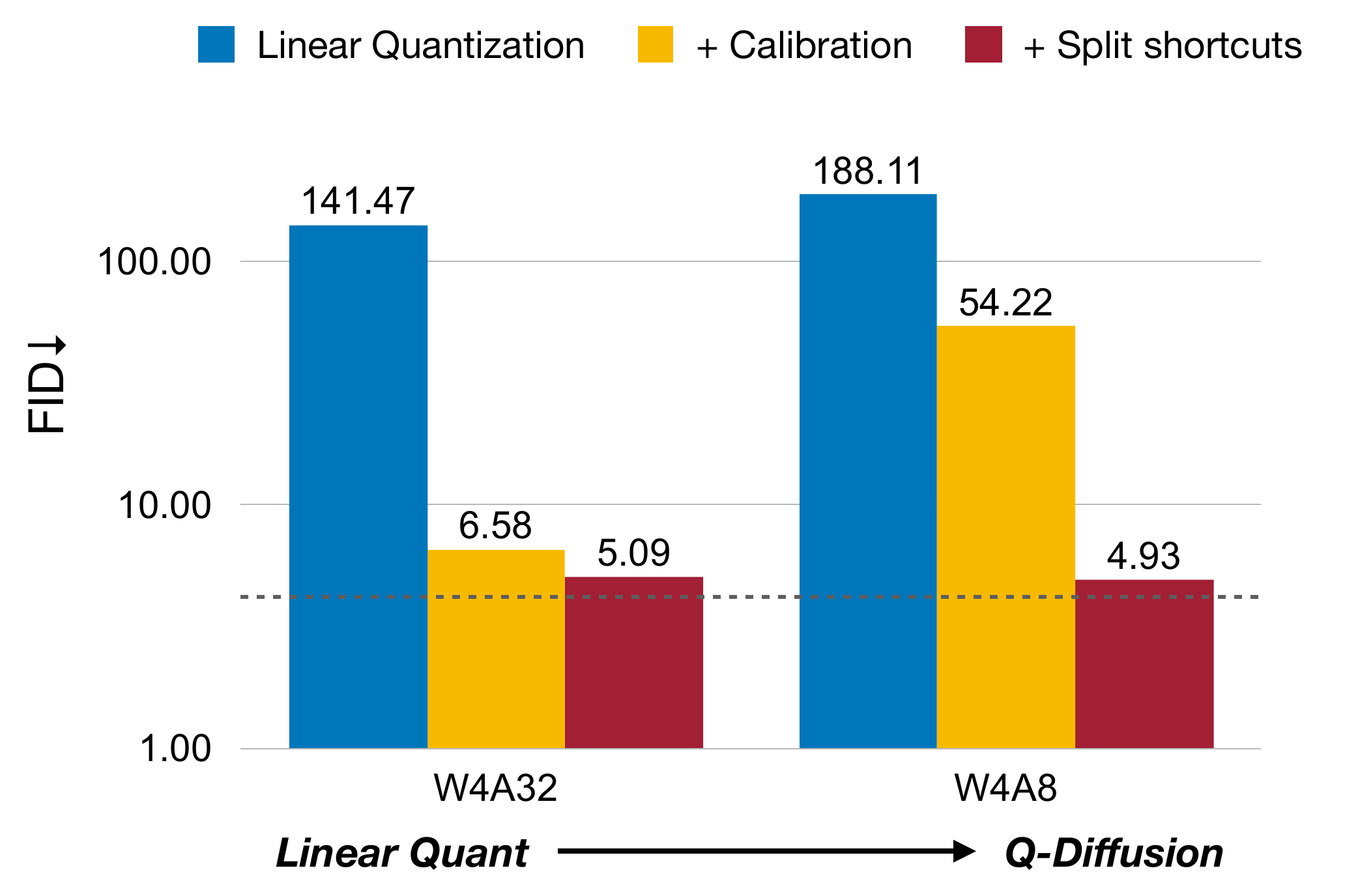}
\vspace{-0.5cm}
\caption{Splitting the shortcut convolution is crucial for both weight and activation quantization. Comparisons on \cifar show that Q-Diffusion could achieve comparable image generation quality to the model with full precision (dashed line) with shortcut splitting.}
\label{abl_spplit}
\vspace{-0.3cm}
\end{figure}

\paragraph{Effects of Split}
Previous linear quantization approaches suffer from severe performance degradation as shown in \fig{abl_spplit}, where 4-bit weight quantization achieves a high FID of 141.47 in DDIM \cifar generation. Employing additional 8-bit activation quantization further degrades the performance (FID: 188.11). By splitting shortcuts in quantization, we significantly improve the generation performance, achieving an FID of 4.93 on W4A8 quantization. 
\vspace{-0.3cm}

\begin{figure}[th]
\centering
\includegraphics[width=1.0\linewidth]{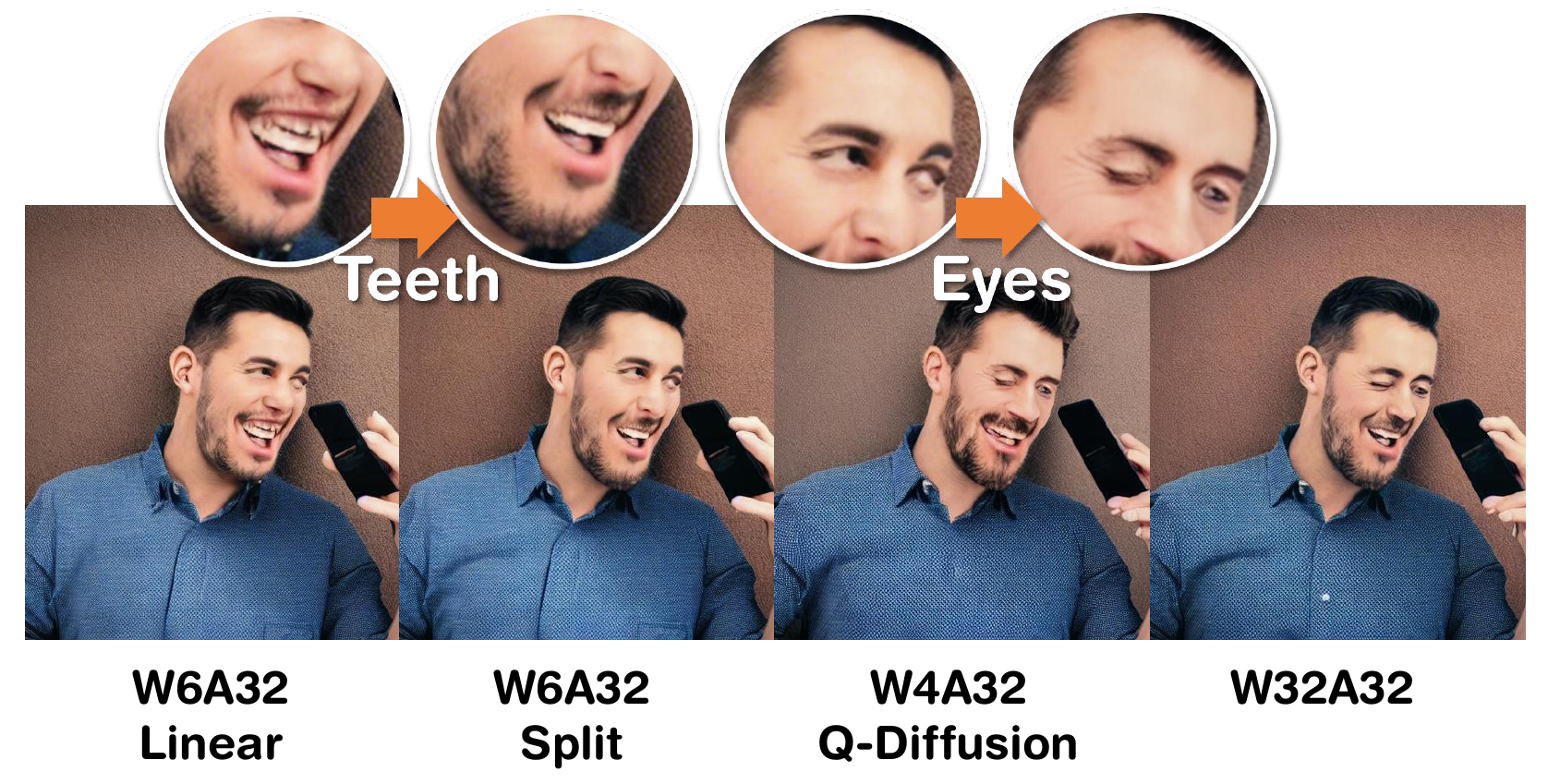}
\vspace{-15pt}
\caption{Examples of text-to-image generation with a quantized Stable Diffusion model. Naive linear quantization degrades the appearance of teeth, which gets fixed by shortcut splitting. \name further improves the semantic consistency of eyes through calibration.}
\label{fig:phoneman}
\vspace{-5pt}
\end{figure}

%% file: 6conclusion.tex
\section{Conclusion}
This work studies the use of quantization to accelerate diffusion models. We propose Q-Diffusion, a novel post-training quantization scheme that conducts calibration with multiple time steps in the denoising process and achieves significant improvements in the performance of the quantized model. Q-Diffusion models under 4-bit quantization achieve comparable results to the full precision models.

\section*{Acknowledgement}
We thank Berkeley Deep Drive, Intel Corporation, Panasonic, and NVIDIA for supporting this research. We also thank Sehoon Kim, Muyang Li, and Minkai Xu for their valuable feedback.

%% file: appendix.tex
\section{Extended Experimental Settings}

\subsection{Implementation Details}
We describe the implementation and compute details of the experiments in this section. We adapt the official implementation for DDIM \cite{song2020denoising}~\footnote{\url{https://github.com/ermongroup/ddim}} and Latent Diffusion \cite{Rombach2021HighResolutionIS}~\footnote{\url{https://github.com/CompVis/latent-diffusion}}. For \sd, we use the CompVis codebase~\footnote{\url{https://github.com/CompVis/stable-diffusion}} and its v1.4 checkpoint. We use the torch-fidelity library~\footnote{\url{https://github.com/toshas/torch-fidelity}} to evaluate FID and IS scores as done in \cite{Rombach2021HighResolutionIS}. We use $100$ denoising time steps for DDIM \cifar. We select $200$ and $500$ denoising time steps for \bed and \church respectively, which are the configurations that achieve the best results provided by~\cite{Rombach2021HighResolutionIS}. For text-guided image generation with \sd, we choose the default PNDM sampler with $50$ time steps.

For quantization experiments, we quantize all weights and activations involved in matrix multiplications, but leave activation functions (e.g. \texttt{SoftMax}, \texttt{SiLU}) and normalization layers (e.g. \texttt{GroupNorm}) running with full precision. Additionally, for Linear Quantization and SQuant experiments, we dynamically update the activation quantizers throughout the image generation process to establish strongest possible baselines, which explains why sometimes their results are better than weight-only quantization cases. For text-guided image generation with \sd, we find that attention matrices in cross attentions are difficult to quantize after the \texttt{SoftMax} and may have considerable influences on the generation quality, so we utilize INT16 mixed-precision for attention scores under W8A8 \& W4A8 cases, while $q$, $k$, $v$ matrices are still quantized down to 8-bit. No special modifications or mixed precision are done for other experiments.

\subsection{Text-guided Image Generation Calibration Dataset Generation Details}
For text-guided image generation with \sd, we need to also include text conditioning in the calibration dataset. We randomly sample text prompts from the MS-COCO dataset, and for each prompt we add a pair of data with both a conditional feature $\vec{c}_t$ and an unconditional feature $\vec{uc}_t$ derived from the prompt. This updated calibration dataset creation process is described by \algo{alg:qdiff_t2i}. Note that we ignore showing the corresponding time embedding $\vec{t}_t$ for each time step $t$ is also added with the sample in Algorithm 1 of the main paper.

\begin{algorithm}
\caption{Q-Diffusion Calibration for Text-guided Image Generation} \label{alg:qdiff_t2i}
\textbf{Require:} Pretrained full precision diffusion model and the quantized diffusion model [$W_\theta$, $\hat{W}_\theta$]  \\
\textbf{Require:} Empty calibration dataset $\mathcal{D}$ \\
\textbf{Require:} Number of denoising sampling steps $T$ \\
\textbf{Require:} Calibration sampling interval $c$, amount of calibration data per sampling step $n$ 
\begin{algorithmic}
\FOR {$t = 1, \ldots, T$ time step}
    \IF{t \% c = 0}
        \STATE Sample $2n$ intermediate inputs $(\vec{x}^{(1)}_t, \vec{c}^{(1)}_t, \vec{t}^{(1)}_t), (\vec{x}^{(1)}_t, \vec{uc}^{(1)}_t, \vec{t}^{(1)}_t), \ldots, (\vec{x}^{(n)}_t, \vec{c}^{(n)}_t, \vec{t}^{(n)}_t), (\vec{x}^{(n)}_t, \vec{uc}^{(n)}_t, \vec{t}^{(n)}_t)$ randomly at $t$ from $W_\theta$ and add them to $\mathcal{D}$
    \ENDIF
\ENDFOR
\FOR {all $i = 1, \ldots, N$ blocks}
    \STATE Update the weight quantizers of the $i$-th block in $\hat{W}_\theta$ with $\mathcal{D}$ and $W_\theta$
\ENDFOR
\IF{do activation quantization}
    \FOR {all $i = 1, \ldots, N$ blocks}
        \STATE Update the activation quantizers step sizes of the $i$-th block with $\hat{W}_\theta$, $W_\theta$, $\mathcal{D}$.
    \ENDFOR
\ENDIF
\end{algorithmic}
\label{alg:calib}
\end{algorithm}

\subsection{Hyperparameters}
Here we provide the hyperparameters used for our Q-Diffusion calibration in \tbl{tab:hyper}.
\begin{table}[ht!]
    \centering
    \begin{tabular}{lllll}
    \toprule
    Experiment  & $T$  & $c$ & $n$ & $N$ \\ 
    \midrule
    DDIM CIFAR-10  & $100$  & $5$ & $256$ & $5120$     \\ 
    LDM-4 LSUN-Bedroom  & $200$  & $10$ & $256$ & $5120$    \\ 
    LDM-8 LSUN-Church  & $500$ & $25$ & $256$ & $5120$      \\
    \sd (weights only)  & $50$  & $2$ & $256 \ (128)$ & $6400$   \\ 
    \sd (weights \& activations)  & $50$  & $1$ & $256 \ (128)$ & $12800$   \\ 
    \bottomrule
    \end{tabular}
    \caption{Hyperparameters for all experiments, including the number of denoising time steps $T$, intervals for sampling calibration data $c$, amount of calibration data per sampling step $n$, and the size of calibration dataset $N$. Note that for \sd with classifier-free guidance, every text prompt ($128$ in total for each sampling step) will add a pair of two samples to the calibration dataset.}
    \label{tab:hyper}
\end{table}

For all unconditional generation experiments, we keep the total calibration dataset size as $5120$ and the amount of calibration data per sampling step as $256$. \name is able to obtain high-quality images with insignificant fidelity loss by uniformly sampling from $20$ time steps without any hyperparameters tuning. For text-guided image generation with \sd, the introduction of text conditioning makes activation quantization harder, thus we sample a larger calibration dataset using all time steps.

\section{Layer-wise Activations Distribution in DDIM and LDM} 
\label{ssec:act_layers}
\begin{figure*}[h]
\centering
\includegraphics[width=\linewidth]{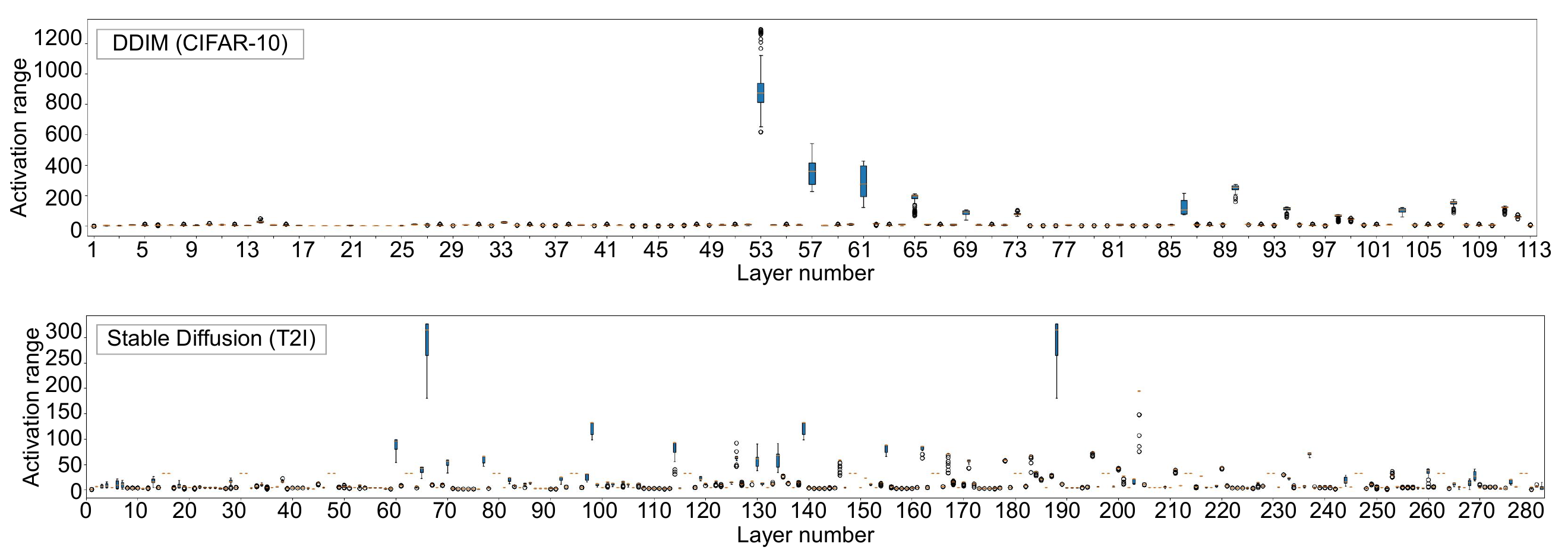}
\includegraphics[width=\linewidth]{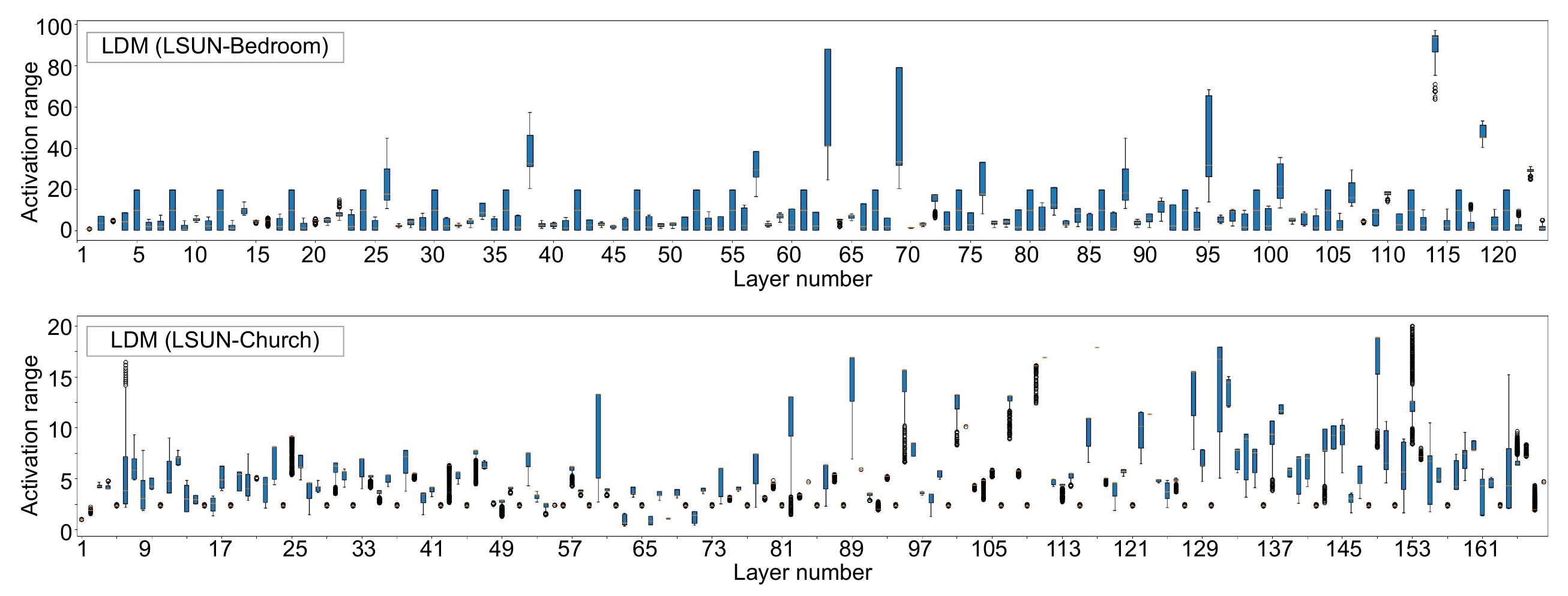}
\caption{Activation ranges of FP32 outputs across layers averaging among all time steps. The figures, from top to bottom, are respectively DDIM, \sd, LDM-Bedroom, and LDM-Church.}
\label{act_layers}
\end{figure*}

We analyze the ranges of activation values across all time steps in DDIM on CIFAR-10, LDM on LSUN-Bedroom and LSUN-Church, and \sd on the text-to-image task. \fig{act_layers} shows that all Conv layers with residual connections in DDIM exhibit noticeably wider activation ranges. Specifically, the first Conv layer can reach up to 1200 and others with residual connections have ranges larger than 100, whereas the majority of the layers without residual connections have ranges less than 50.
Similar results could be observed from \sd with the text-to-image generation task with COCO captions as well as LSUN-Bedroom in latent diffusion. On the other hand, all layers in LDM on LSUN-Church share relatively uniform activation distributions, with ranges $<15$.

\begin{figure*}[h]
\centering
\includegraphics[width=0.85\linewidth]{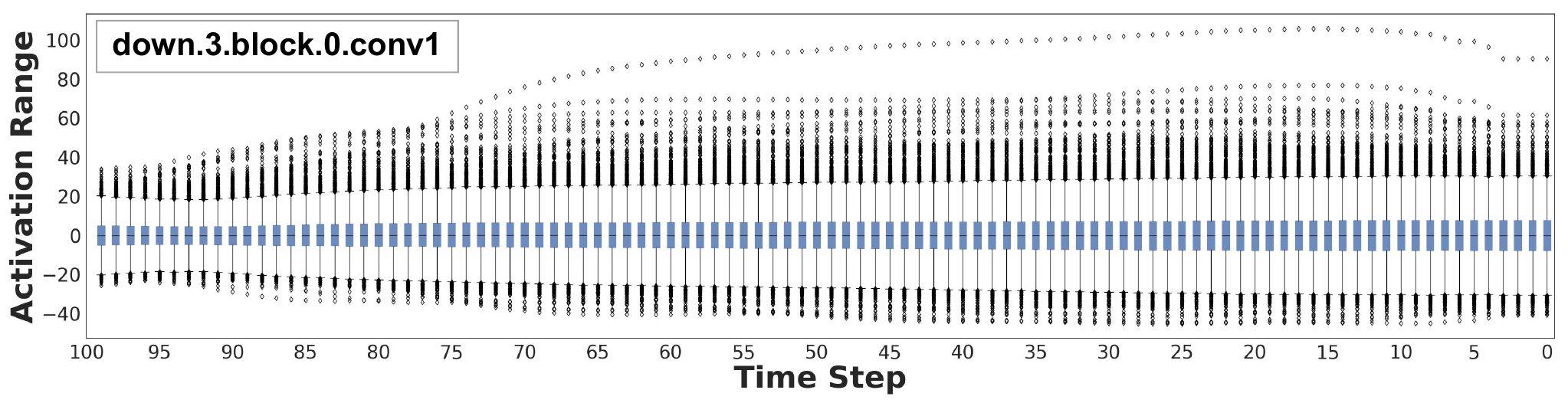}
\includegraphics[width=0.85\linewidth]{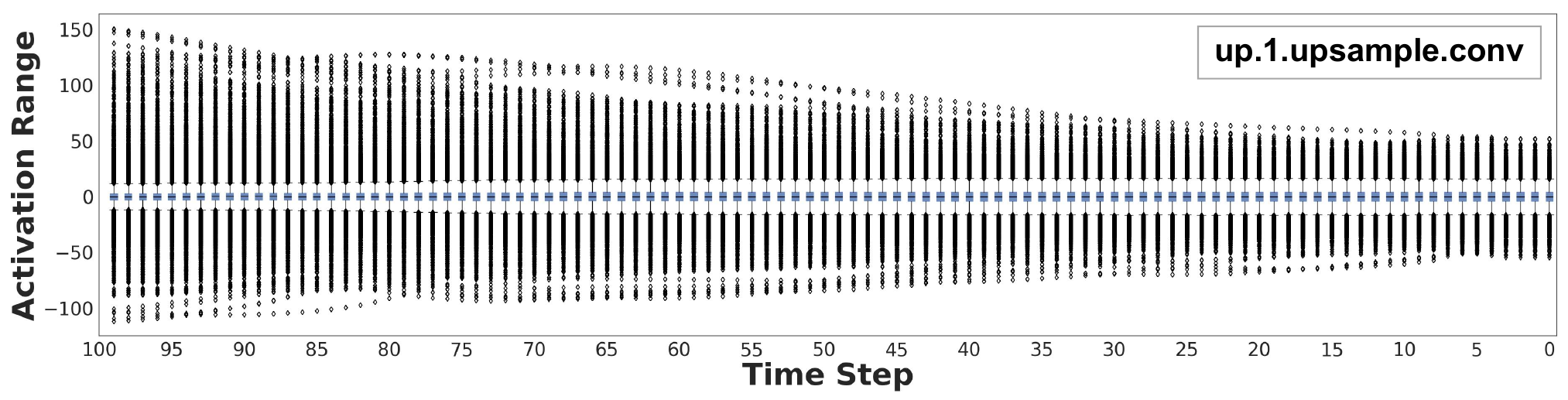}
\includegraphics[width=0.85\linewidth]{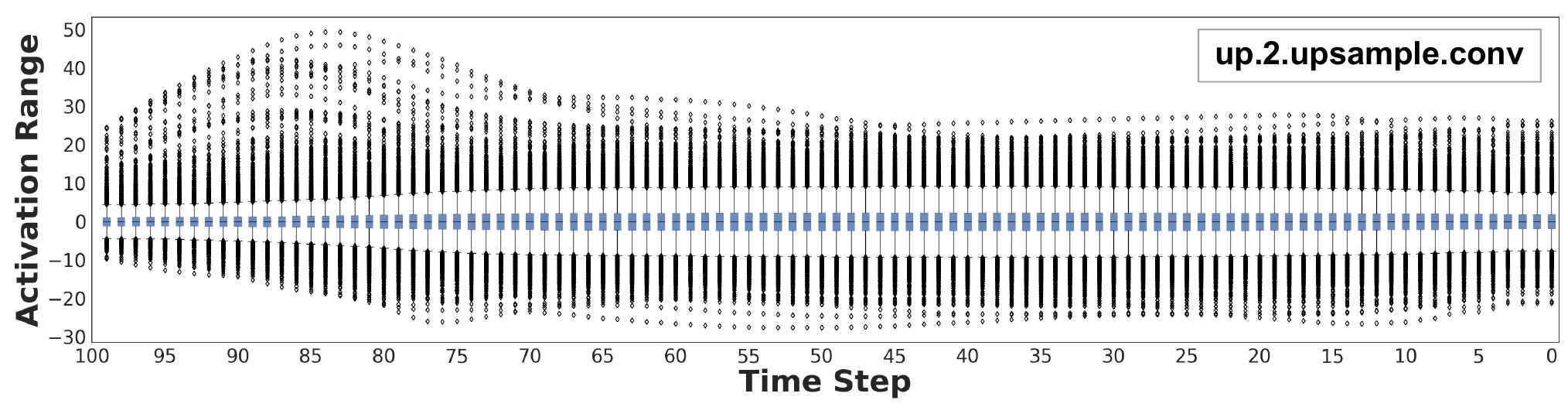}
\includegraphics[width=0.85\linewidth]{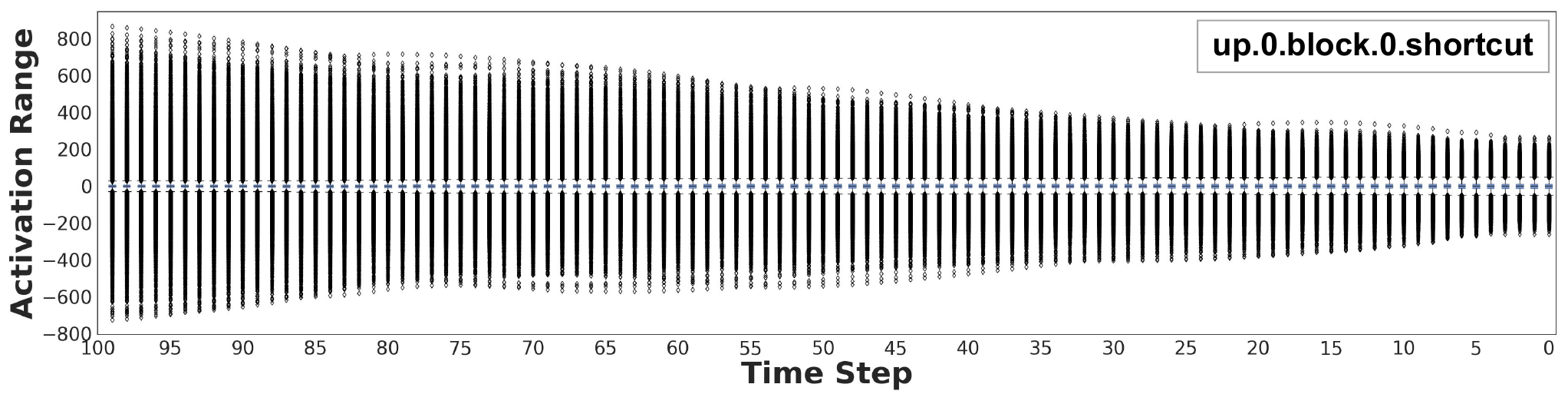}
\caption{The distribution of activation values of multiple layers in DDIM on CIFAR-10 varies significantly across different time steps.}
\label{act_layer_over_step}
\end{figure*}

Furthermore, \fig{act_layer_over_step} illustrates that the distribution of activation values of multiple layers in DDIM on CIFAR-10 varies significantly across different time steps.

\section{Quantitative Evaluation on Text-guided Image Generation}
\label{sec:sd_score}
\begin{figure*}[!ht]
  \begin{center}
  \vspace{-10pt}
  \begin{minipage}[c]{0.49\linewidth}
    \centering
    \vspace{0.1cm}
    \includegraphics[width=0.99\textwidth]{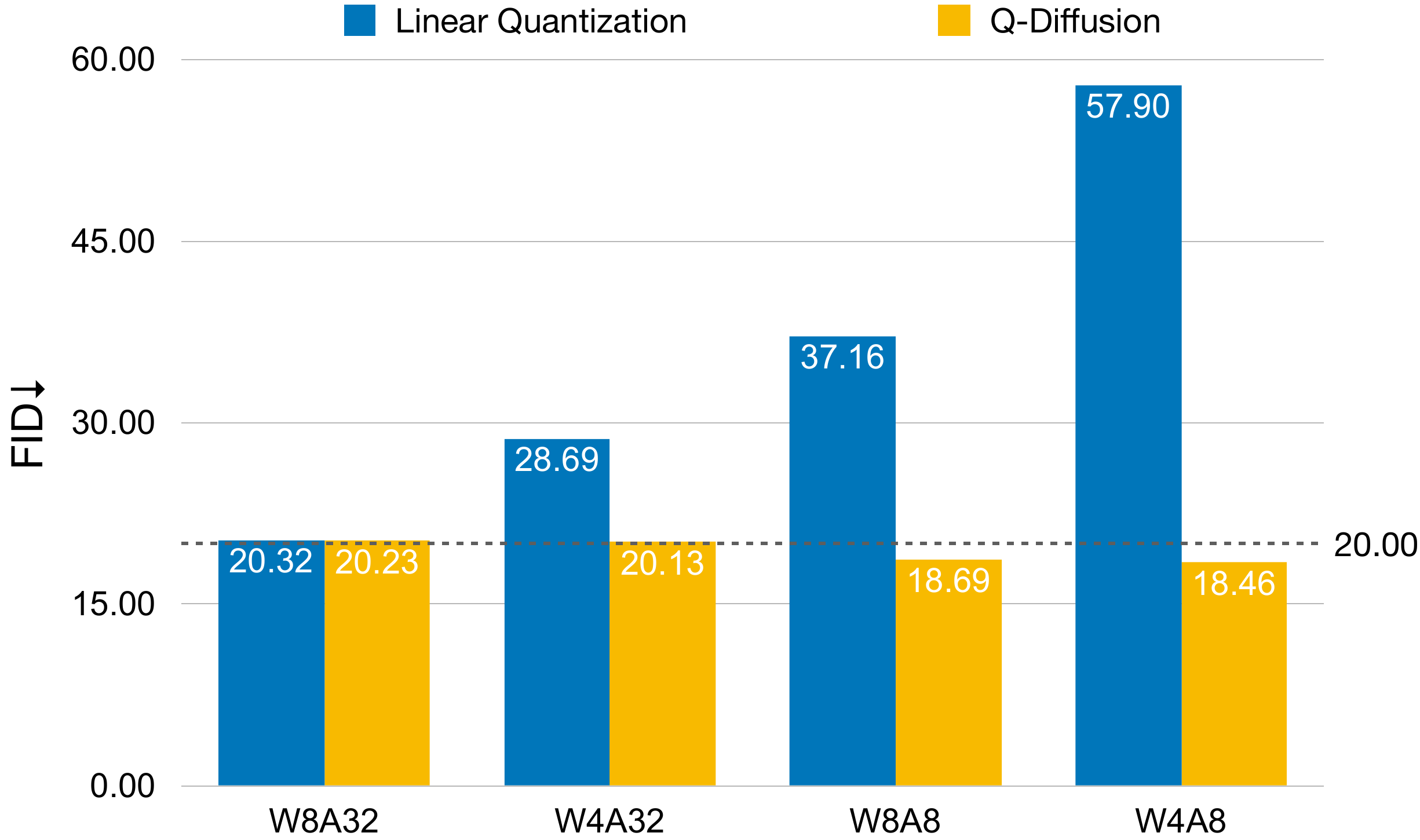}
    \footnotesize (a) \name has negligible increases in FID \\
  \end{minipage}\hfill
  \begin{minipage}[c]{0.49\linewidth}
    \centering
    \vspace{0.1cm}
    \includegraphics[width=0.99\textwidth]{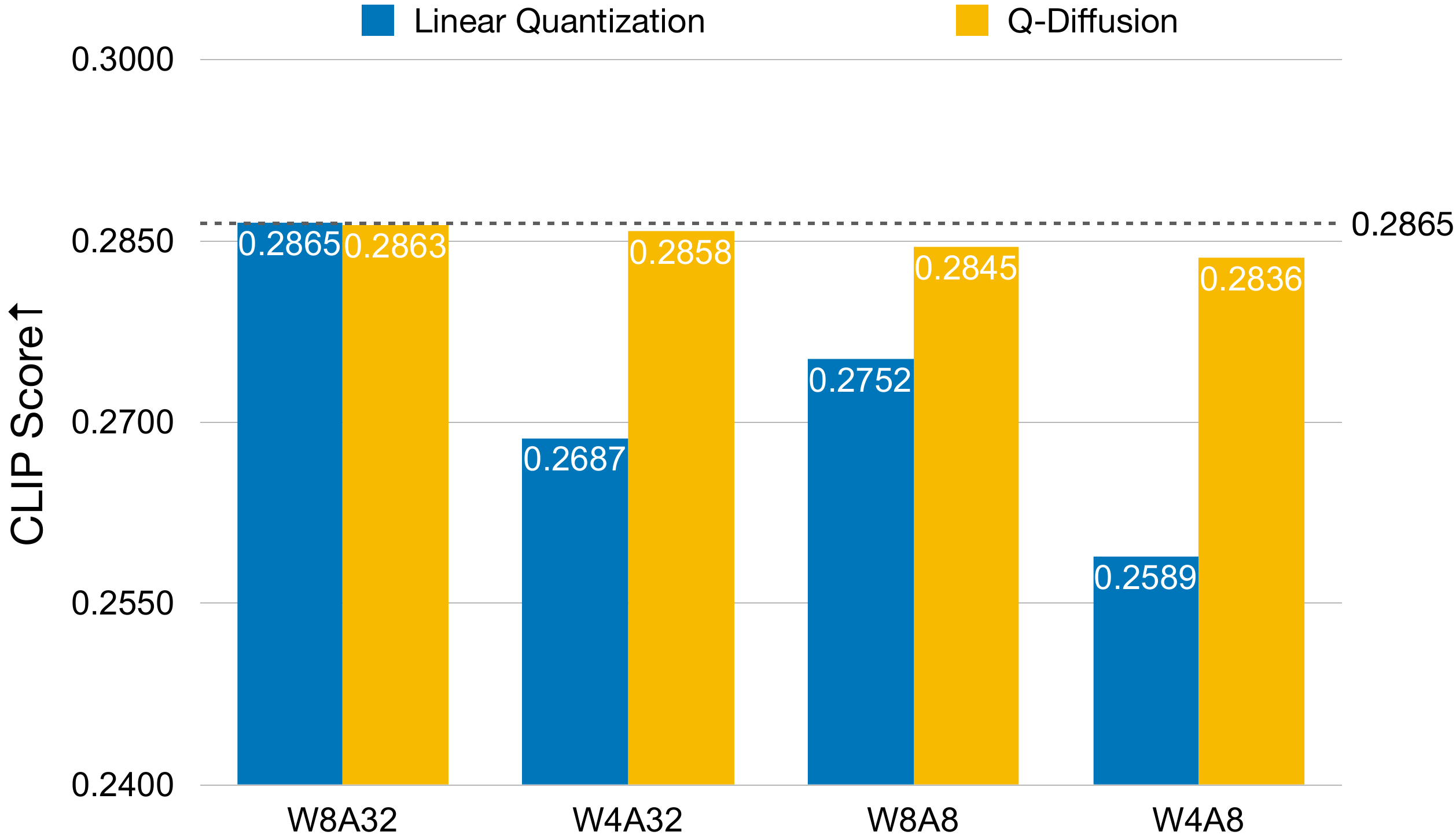}
    \footnotesize (b) \name maintains similar CLIP score \\
  \end{minipage}
  \vspace{-0.15cm}
  \caption{\sd (cfg scale = 7.5) 512 $\times$ 512 text-guided image synthesis FID and CLIP score results quantized using Q-Diffusion and Linear Quantization under different precisions. The dotted lines values are obtained under full precision.}
  \label{fig:sd_coco}
  \end{center}
\end{figure*}

To quantitatively evaluate the extent of the impacts on generation performance induced by quantization, we follow the practice in~\cite{Rombach2021HighResolutionIS}, Stable Diffusion v1-5 model card~\footnote{\url{https://huggingface.co/runwayml/stable-diffusion-v1-5}}, and Diffusers library~\footnote{\url{https://huggingface.co/docs/diffusers/main/en/conceptual/evaluation}} to generate $10$k images using prompts from the MS-COCO~\cite{Lin2014MicrosoftCC} 2017-val dataset. Subsequently, we compute the FID~\cite{heusel2017fid} and CLIP score~\cite{hessel2021clipscore} against the 2017-val dataset. The ViT-B/16 is used as the backbone when computing the CLIP score. Results are illustrated in \fig{fig:sd_coco}.

Our \name has minimal quality degradation in generated images measured by these two metrics under all settings, while the direct Linear Quantization incurs significant quality drops, especially when the activations are also quantized. Note that FID and CLIP score on COCO may not be good metrics that align well with human preferences; we do observe that slight artifacts appear more often on images generated with models that have both weights and activations quantized by \name, while these are not reflected by the FID results.

\section{Study of Combining with Fast Samplers}
\begin{figure*}[h]
  \begin{center}
  \begin{minipage}[c]{0.99\linewidth}
    \centering
    \includegraphics[width=0.99\textwidth]{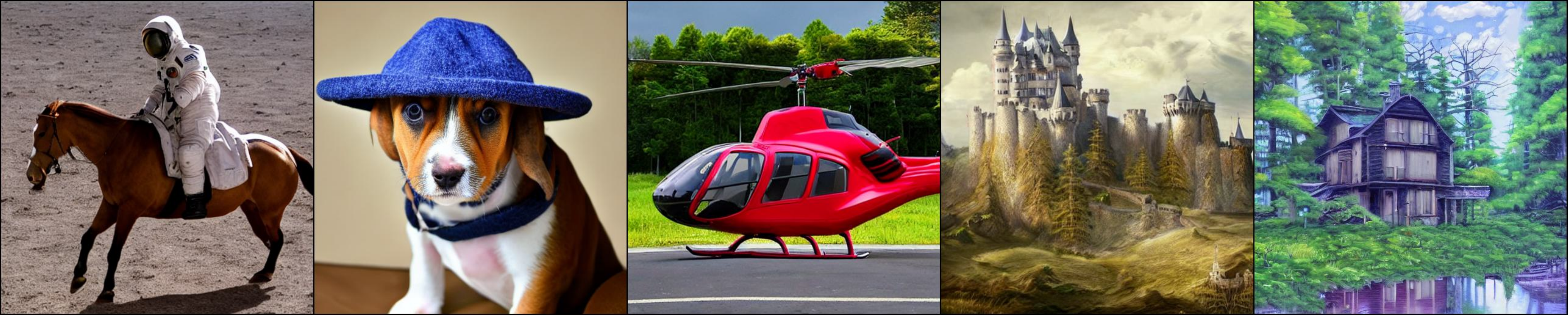}
    \vspace{0.3cm}
    \small \name (W4A32)
  \end{minipage}\hfill
  \begin{minipage}[c]{0.99\linewidth}
    \centering
    \includegraphics[width=0.99\textwidth]{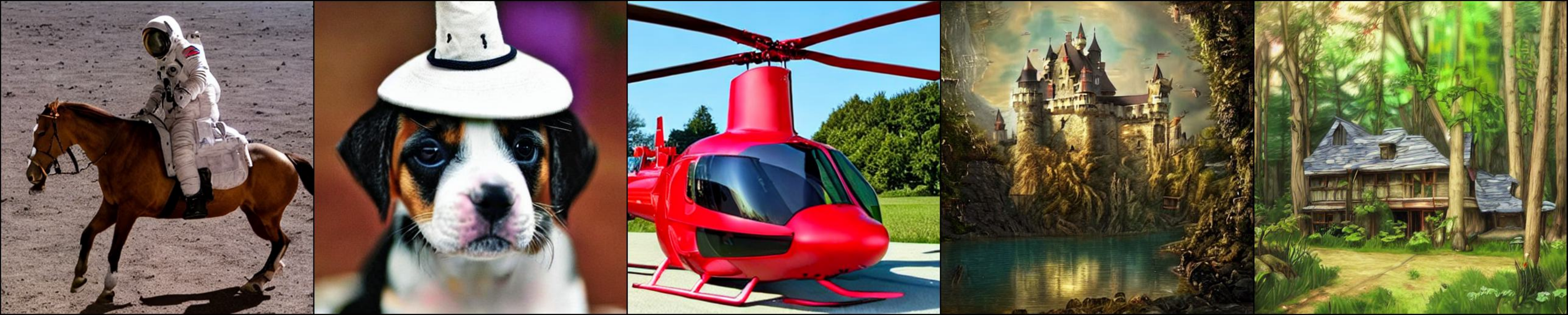}
    \vspace{0.3cm}
    \small \name DPM-Solver++ (W4A32)
  \end{minipage}
  \caption{Text-guided image generation results on 512 $\times$ 512 resolution from our INT4 weights-quantized \sd with default PNDM 50 time steps and DPM-Solver++ 20 time steps.}
  \label{fig:dpm}
  \end{center}
\end{figure*}

Another line of work to speed-up diffusion models is to find shorter and more effective sampling trajectories in order to reduce the number of steps in the denoising process. These approaches tackle an orthogonal factor that \name is addressing, indicating that there's great potential to design a method to take advantage of both directions. Here we investigate if \name can be combined with DPM-Solver~\cite{Lu2022DPMSolverAF, Lu2022DPMSolverFS}, a fast high-order solver for diffusion ODEs that can greatly bring down the number of steps required for generation. For unconditional generation, we use a 3rd-order DPM-Solver++ as suggested by the authors, and sample for 50 time steps, which is the number of steps required to get a converged sample. For text-guided image generation with \sd, we use 2nd-order DPM-Solver++ with 20 time steps. We directly apply this DPM-Solver++ sampler to our INT$4$ quantized model. Results are shown in \tbl{tab:dpm-solver} and \fig{fig:dpm}. 

\name only has a minor performance drop when only weights are quantized. The generation quality degrades under W4A8 precision, but all \name results still outperform Linear Quant and SQuant with $100$, $200$, and $500$ steps for \cifar, \bed, and \church respectively. The reason is likely due to the distribution of activations becoming inconsistent with how \name is calibrated when the sampling trajectories change. We leave the design for a systematic pipeline that can effectively combine these two directions in diffusion model acceleration as future work.

\section{Comparing with PTQ4DM~\cite{shang2022posttraining}}
\label{sec:ptq4dm}
We evaluated Q-Diffusion on the settings employed in~\cite{shang2022posttraining}, which computed Inception Score (IS), Frechet Inception Distance (FID), and sFID~\cite{nash2021generating} over only 10k generated samples. Although~\cite{shang2022posttraining} did not specify details in the paper, their official implementation computed activation-to-activation matrix multiplications in the attention ($q * k$ and $attn * v$) in FP16/32\footnote{\url{https://github.com/42Shawn/PTQ4DM} (05/31/2023)}, while we conducted them in full-integer. These matmuls account for a substantial portion of FLOPs (e.g. 9.8\% of the model in SD) and can induce considerable memory overheads~\cite{dao2022flashattention}, which subsequently increase the inference costs. Contrarily, our work reduces the memory \& compute in this part by 2x/4x theoretically.

The evaluation results are demonstrated in \tbl{tab:compare}, where numbers inside the parentheses of PTQ4DM are its results reproduced with integer attention matmuls. \name consistently outperforms PTQ4DM~\cite{shang2022posttraining}, which achieves better results with attention matmuls in INT8 than PTQ4DM with them computed in FP16/32. Note that directly applying~\cite{shang2022posttraining} to quantize attention matmuls in 8-bit would even further degrade generation quality, as shown by the numbers inside parentheses.

\begin{table}[!hbt]
    \centering
    \vspace{-9pt}
    \caption{Q-Diffusion and PTQ4DM~\cite{shang2022posttraining} results. The numbers inside the PTQ4DM parentheses refer to~\cite {shang2022posttraining} results with INT8 attention act-to-act matmuls.}
    \label{tab:compare}
    \begin{tabular}{ccccc}
        \toprule
        Task & Method & IS$\uparrow$ & FID$\downarrow$ & sFID$\downarrow$ \\ \midrule
        \multirow{4}{*}{\shortstack{\cifar\\ DDIM 100 steps}} & FP & 9.18 & 10.05 & 19.71 \\ 
        & PTQ4DM (W8A8) & 9.31 (9.02) & 14.18 (19.59) & 22.59 (20.89) \\ 
        & Q-Diffusion (W8A8) & \textbf{9.47} & \textbf{7.82} & \textbf{17.96} \\ 
        & Q-Diffusion (W4A8) & 9.19 & 8.85 & 19.64 \\ 
        \midrule
        \multirow{4}{*}{\shortstack{\cifar\\ DDIM 250 steps}} & FP & 9.19 & 8.83 & 18.31 \\ 
        & PTQ4DM (W8A8) & \textbf{9.70} (9.30) & 11.66 (16.54) & 19.71 (20.08) \\ 
        & Q-Diffusion (W8A8) & 9.60 & \textbf{8.00} & \textbf{18.13} \\ 
        & Q-Diffusion (W4A8) & 9.18 & 8.54 & 18.58 \\ 
        \bottomrule
    \end{tabular}
\end{table}

\begin{table}[h]
    \centering
    \caption{\name results when directly applying 3rd-order DPM-Solver++ with $50$ denoising time steps.}
    \vspace{-5pt}
\label{tab:dpm-solver}
    \begin{tabular}{ccc}
    \toprule
     Task & Bits (W/A) & FID$\downarrow$  \\
    \midrule
    DDIM \cifar & 32/32 & 3.57 \\
    DDIM \cifar & 4/32 & 5.38 \\
    DDIM \cifar & 4/8 & 10.27 \\
    \midrule
    LDM-4 \bed & 32/32 & 4.27 \\
    LDM-4 \bed  & 4/32 & 4.88 \\
    LDM-4 \bed  & 4/8 & 10.77 \\
    \midrule
    LDM-8 \church  & 32/32 & 5.40 \\
    LDM-8 \church & 4/32 & 5.74 \\
    LDM-8 \church & 4/8 & 8.19 \\
    \bottomrule
    \end{tabular}
\end{table}

\section{Limitations of this work}
\label{sec:limit}

This work focuses on providing a PTQ solution for the noise estimation network of the diffusion models on the unconditional image generation task. Meanwhile, we notice the recent advancement of text-guided image generation~\cite{Rombach2021HighResolutionIS} and other multi-modality conditional generation tasks. As we have demonstrated the possibility of directly applying Q-Diffusion to the noise estimation network of \sd, we believe it is important to provide a systematic analysis of the quantization's impact on the text encoder and the cross-attention mechanism for the classifier-free guidance conditioning, to enable a fully quantized conditional generation framework.
For unconditional generation, this work discovers the need to sample calibration data across all time steps, and apply specialized split quantizers for the concatenation layers in the noise estimation model. The combination of these techniques demonstrates good performance for quantized diffusion models. Meanwhile, there exist other interesting design choices, like non-uniform sampling across different time steps, and additional quantizer design for attention softmax output, \etc, that can be explored.
We leave further investigation of these points as future work.

\subsection{Non-uniform sampling methods that Did Not Work}
As a preliminary exploration of non-uniform calibration data sampling across time steps, we explore the following 3 sampling methods. Yet none of those achieves better performance than Uniform sampling as proposed in this paper under the same amount of calibration data (5120), as shown in Table~\ref{tab:nonuniform}. 
\paragraph{Std}
Since we observe the diverse activation range across time steps in Fig. 5, we would like to sample more data from the time step with a larger variance in its distribution, so as to better represent the overall output distribution across all time steps. To this end, we propose to sample calibration data from each time step in proportion to the pixel-wise standard deviation (Std) of each time step. Specifically, we randomly sample $256$ $x_t$ among all time steps and compute the Std of all pixel values in $x_t$ at each time step, which we denote as $s_t$. Then for calibration data, we sample $\frac{s_t}{\sum_t s_t}N$ examples out of the total $N$ calibration data from time step $t$. 

\paragraph{Norm Std}
Similar to Std, we also consider modeling the variance of each time step's distribution with the standard deviation of $||x_t||_2$, instead of the Std of all pixel values. We expect the Norm Std can better capture the diversity across different samples instead of capturing the pixel-wise diversity within each sample compared to pixel-wise Std.

\paragraph{Unsupervised Selective Labeling (USL)}
We also try to use Unsupervised Selecting Labeling \cite{wang2022unsupervised} to select both representative and diverse samples as the calibration samples. The intuition is that samples that are both representative and diverse could provide a wide range of activations that we will encounter at inference time, focusing on which could bring us good performance on generation most of the time. We select 5120 samples in total for CIFAR-10 by combining the samples for all time steps. We adopt the training-free version of Unsupervised Selective Labeling for sample selection, with the pooled latent space feature from the noise estimation UNet as the selection feature. 

\begin{table}[h]
    \centering
    \caption{Quantization results for unconditional image generation with DDIM on CIFAR-10 (32 $\times$ 32). We compare different calibration data sampling schemes under W4A32 quantization.}
    \vspace{-5pt}
\label{tab:nonuniform}
    \begin{tabular}{ccccc}
    \toprule
     Method & Std & Norm Std & USL &\ourcell Uniform (ours)  \\
    \midrule
    FID$\downarrow$ & 5.66 & 5.58 & 5.54 &\ourcell \textbf{5.09} \\
    \bottomrule
    \end{tabular}
\vspace{-12pt}
\end{table}

\section{Additional Random Samples}
\label{sec:add}

In this section, we provide more random samples from our weight-only quantized and fully quantized diffusion models obtained using \name and Linear Quantization under 4-bit quantization. Results are shown in the figures below.

\vspace{12pt}

\begin{figure*}[h]
  \begin{center}
  \begin{minipage}[c]{0.24\linewidth}
    \centering
    \includegraphics[width=0.99\textwidth]{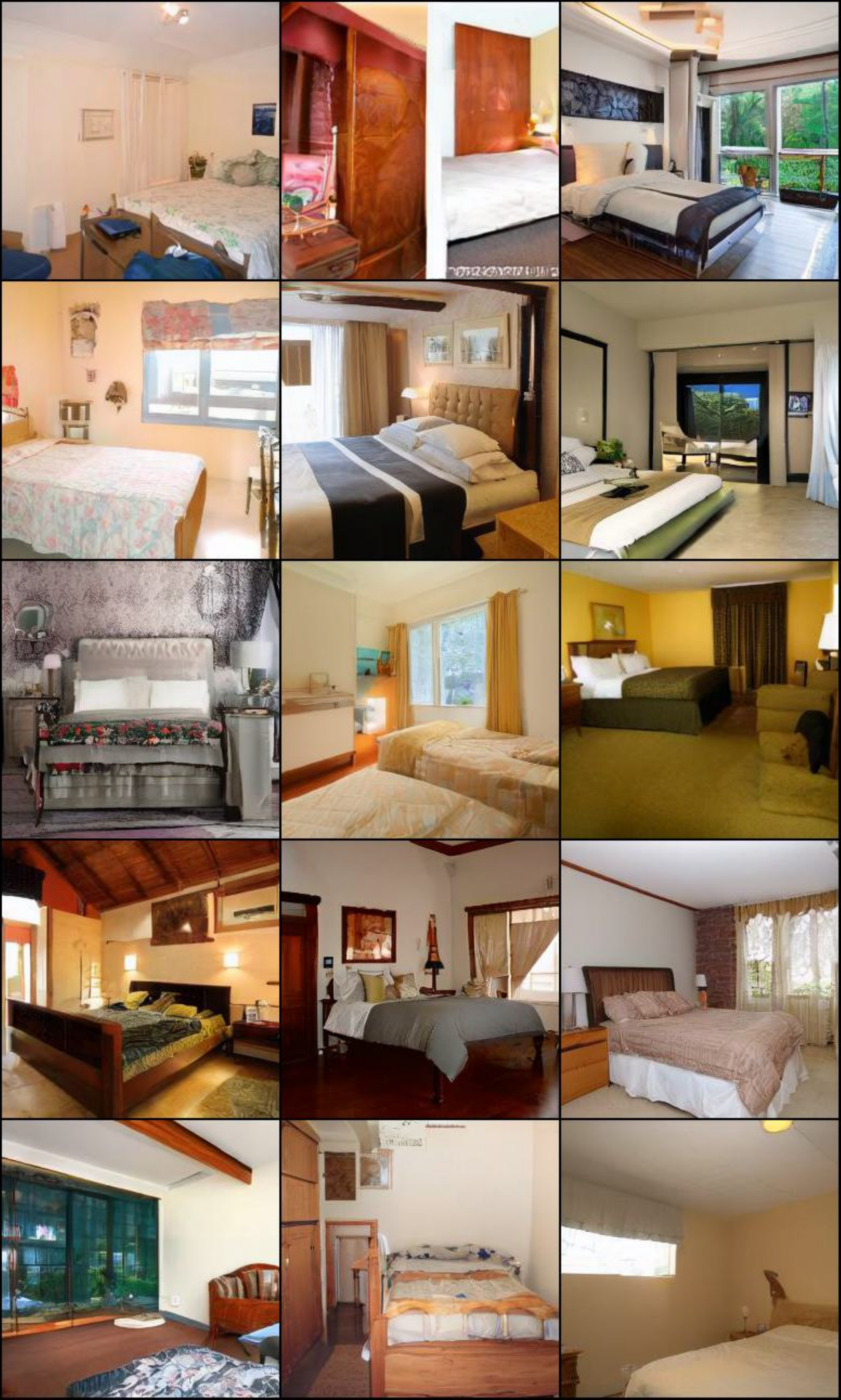}
    \vspace{0.3cm}
    \small \name (W4A32)
  \end{minipage}\hfill
  \begin{minipage}[c]{0.24\linewidth}
    \centering
    \includegraphics[width=0.99\textwidth]{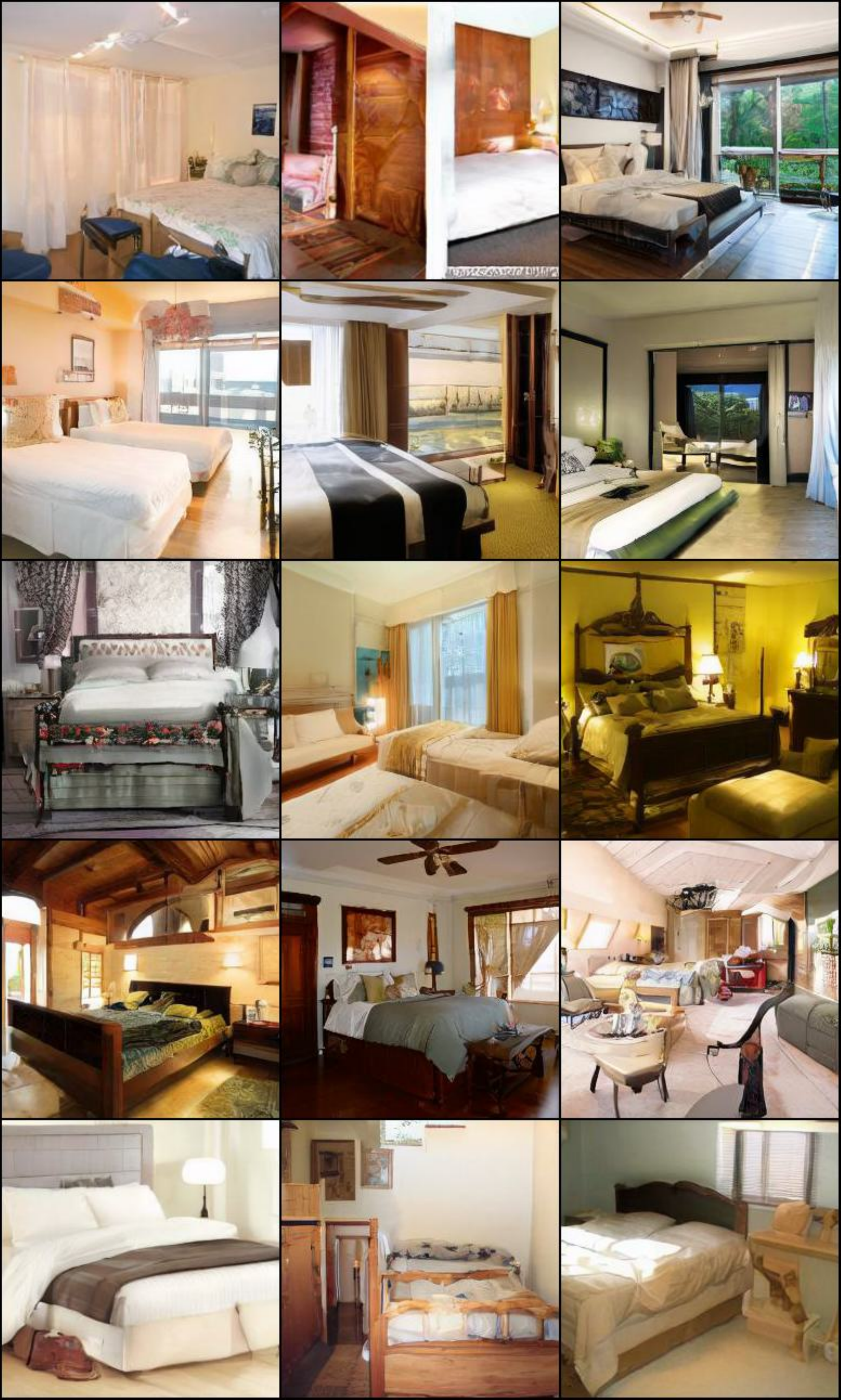}
    \vspace{0.3cm}
    \small \name (W4A8)
  \end{minipage}\hfill
  \begin{minipage}[c]{0.24\linewidth}
    \centering
    \includegraphics[width=0.99\textwidth]{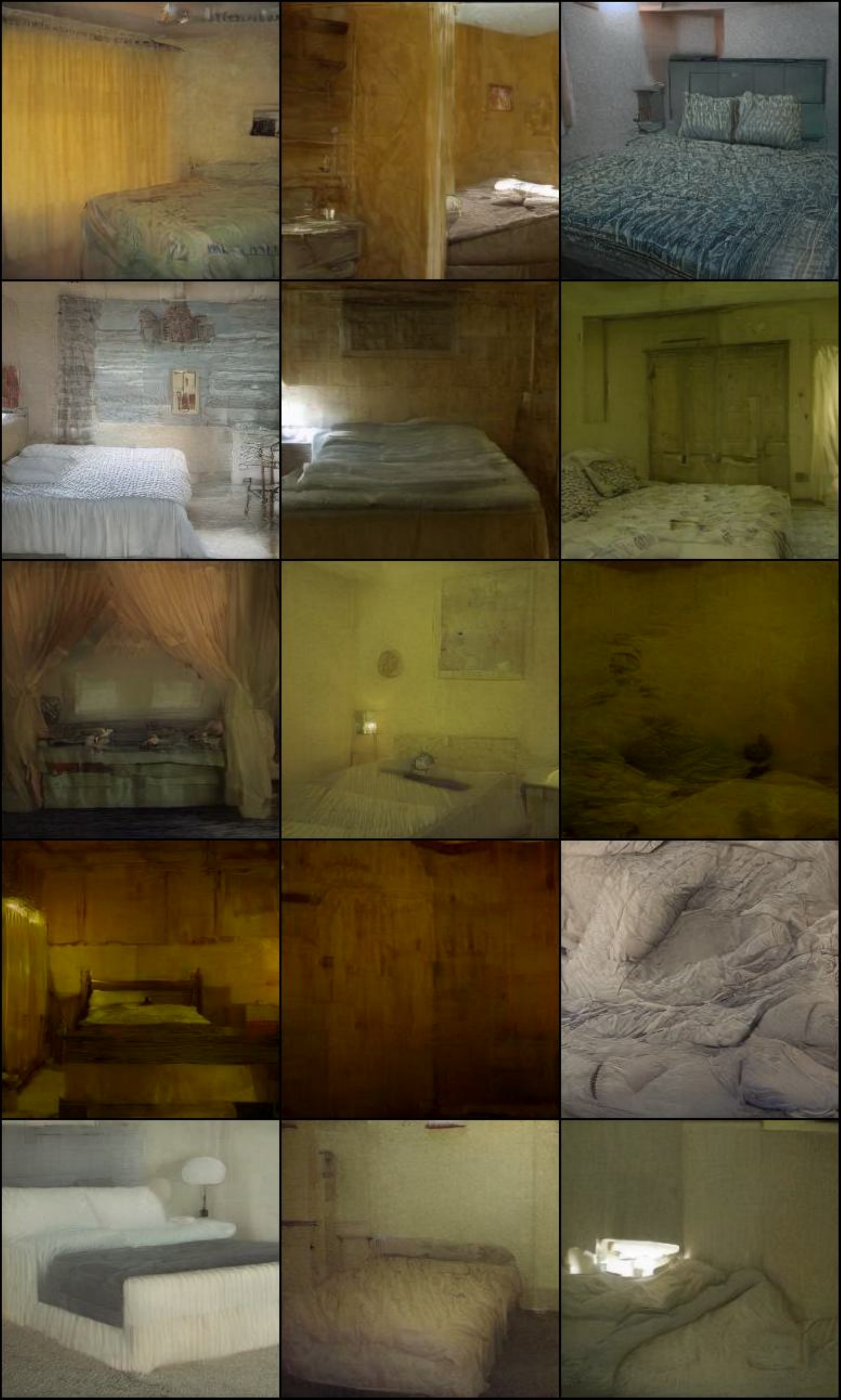}
    \vspace{0.3cm}
    \small Linear Quant (W4A32)
  \end{minipage}\hfill
  \begin{minipage}[c]{0.24\linewidth}
    \centering
    \includegraphics[width=0.99\textwidth]{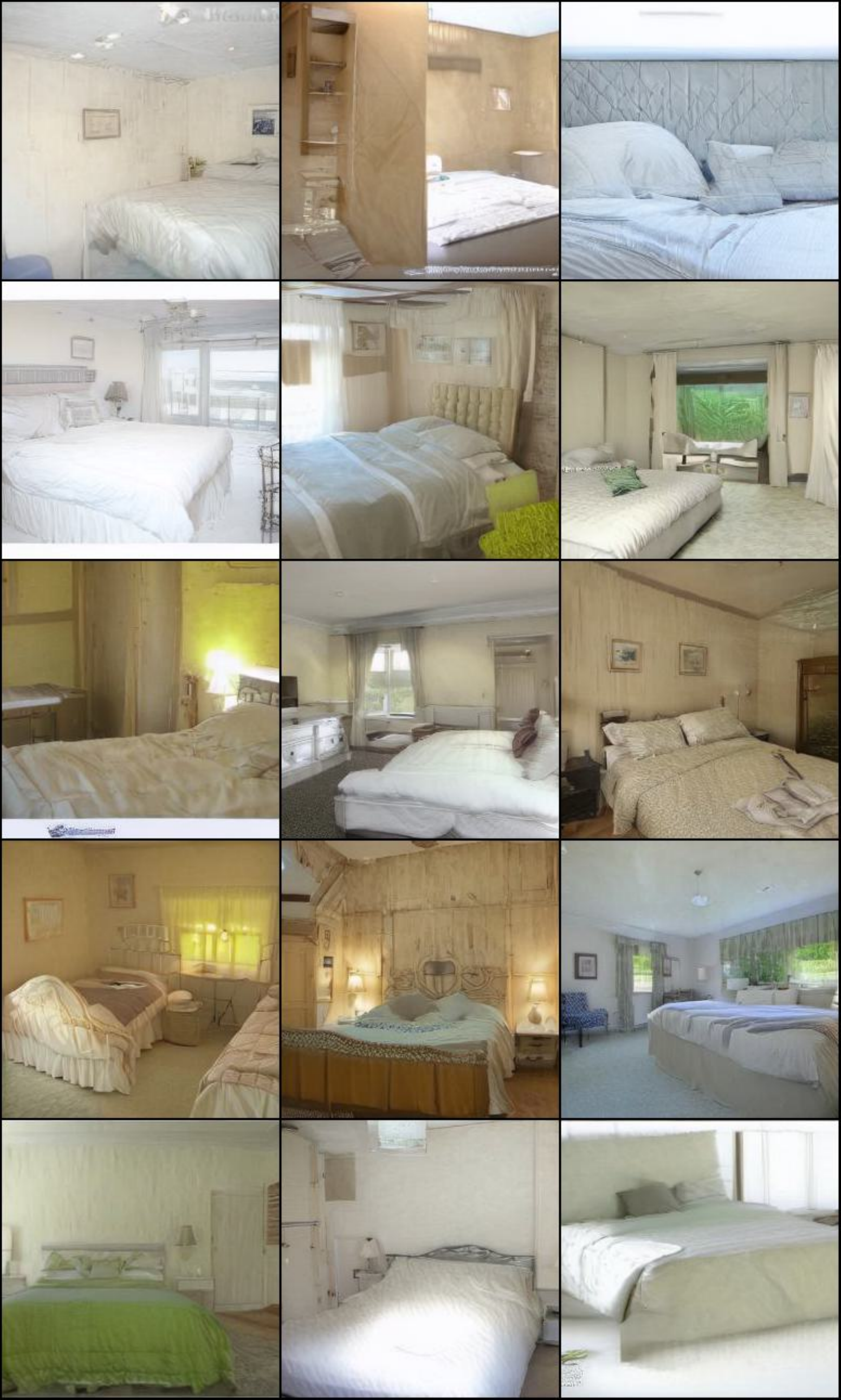}
    \vspace{0.3cm}
    \small Linear Quant (W4A8)
  \end{minipage}
  \vspace{-0.2cm}
  \caption{Random samples from our INT4 quantized 256 $\times 256$ LSUN-Bedroom models with a fixed 
 random seed.}
  \end{center}
\end{figure*}

\vspace{-0.4cm}

\begin{figure*}[h]
  \begin{center}
  \begin{minipage}[c]{0.24\linewidth}
    \centering
    \includegraphics[width=0.99\textwidth]{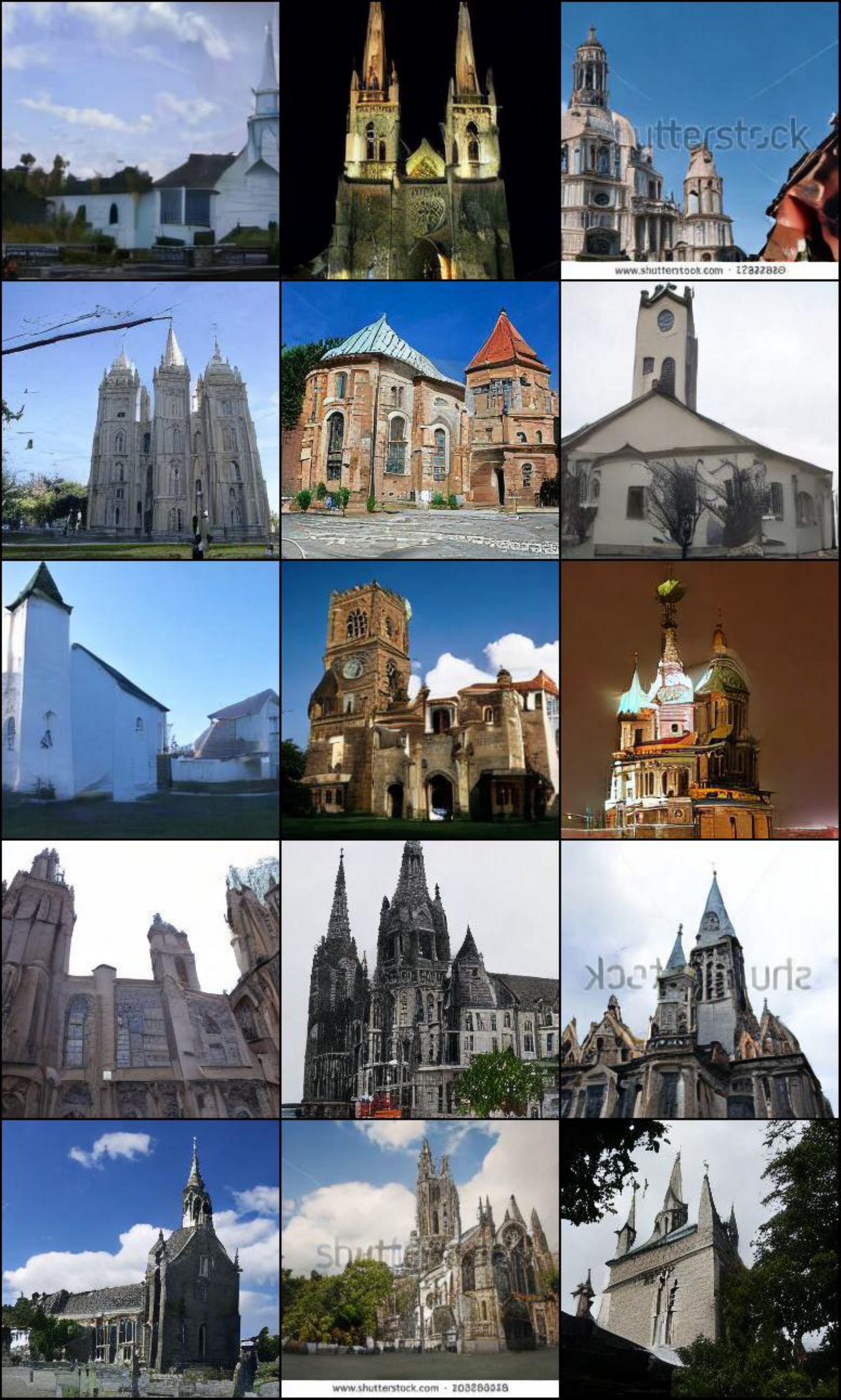}
    \vspace{0.3cm}
    \small \name (W4A32)
  \end{minipage}\hfill
  \begin{minipage}[c]{0.24\linewidth}
    \centering
    \includegraphics[width=0.99\textwidth]{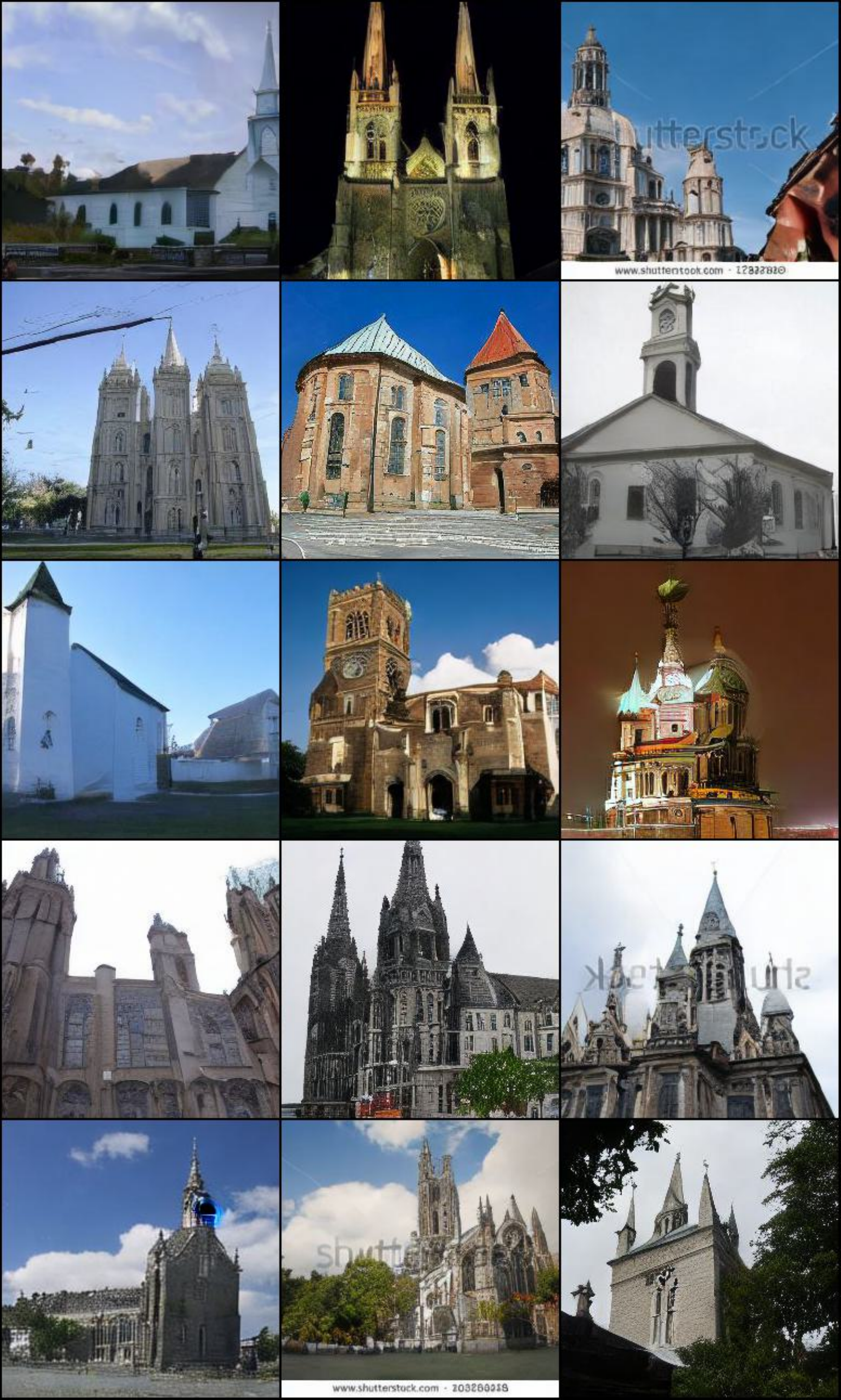}
    \vspace{0.3cm}
    \small \name (W4A8)
  \end{minipage}\hfill
  \begin{minipage}[c]{0.24\linewidth}
    \centering
    \includegraphics[width=0.99\textwidth]{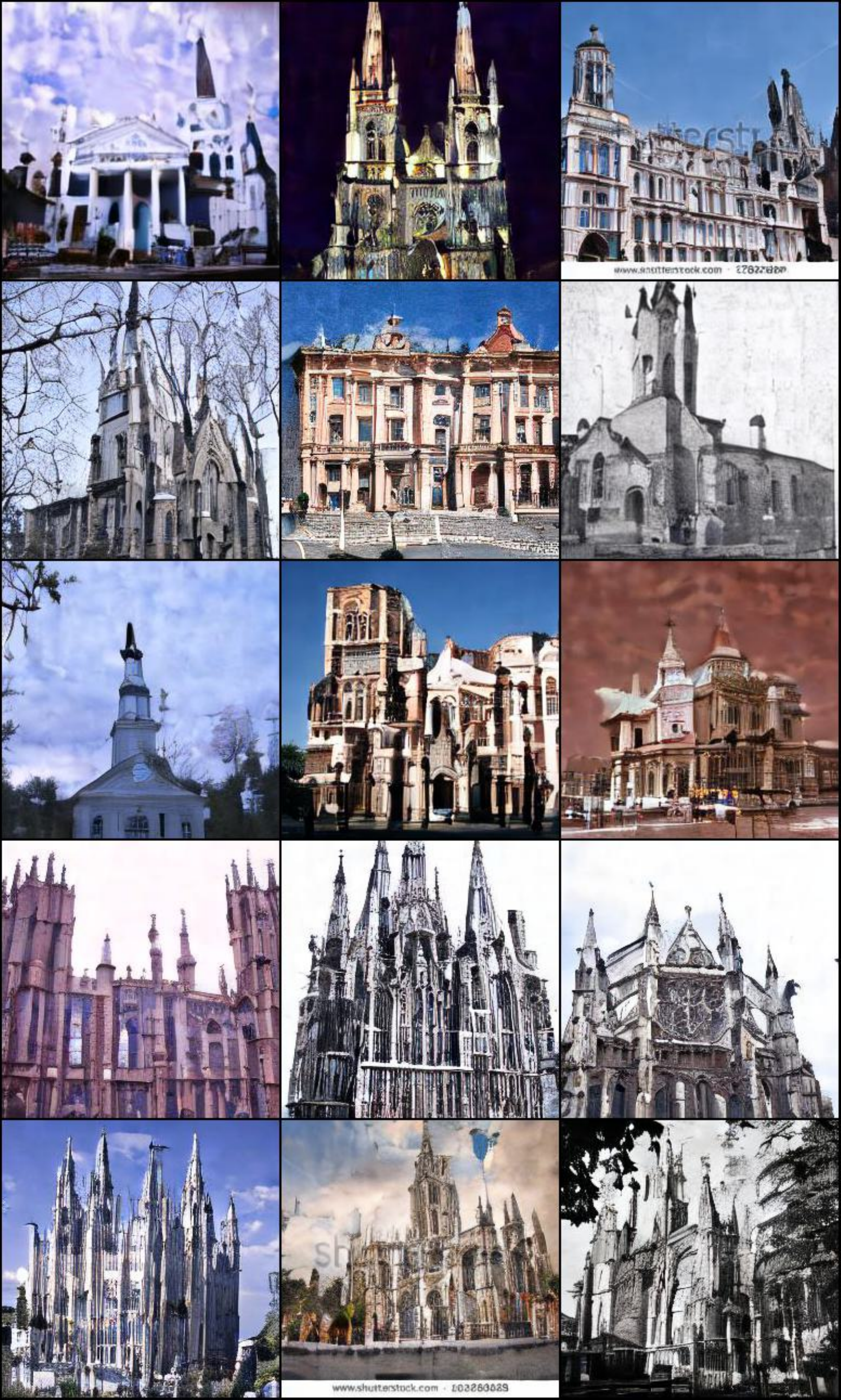}
    \vspace{0.3cm}
    \small Linear Quant (W4A32)
  \end{minipage}\hfill
  \begin{minipage}[c]{0.24\linewidth}
    \centering
    \includegraphics[width=0.99\textwidth]{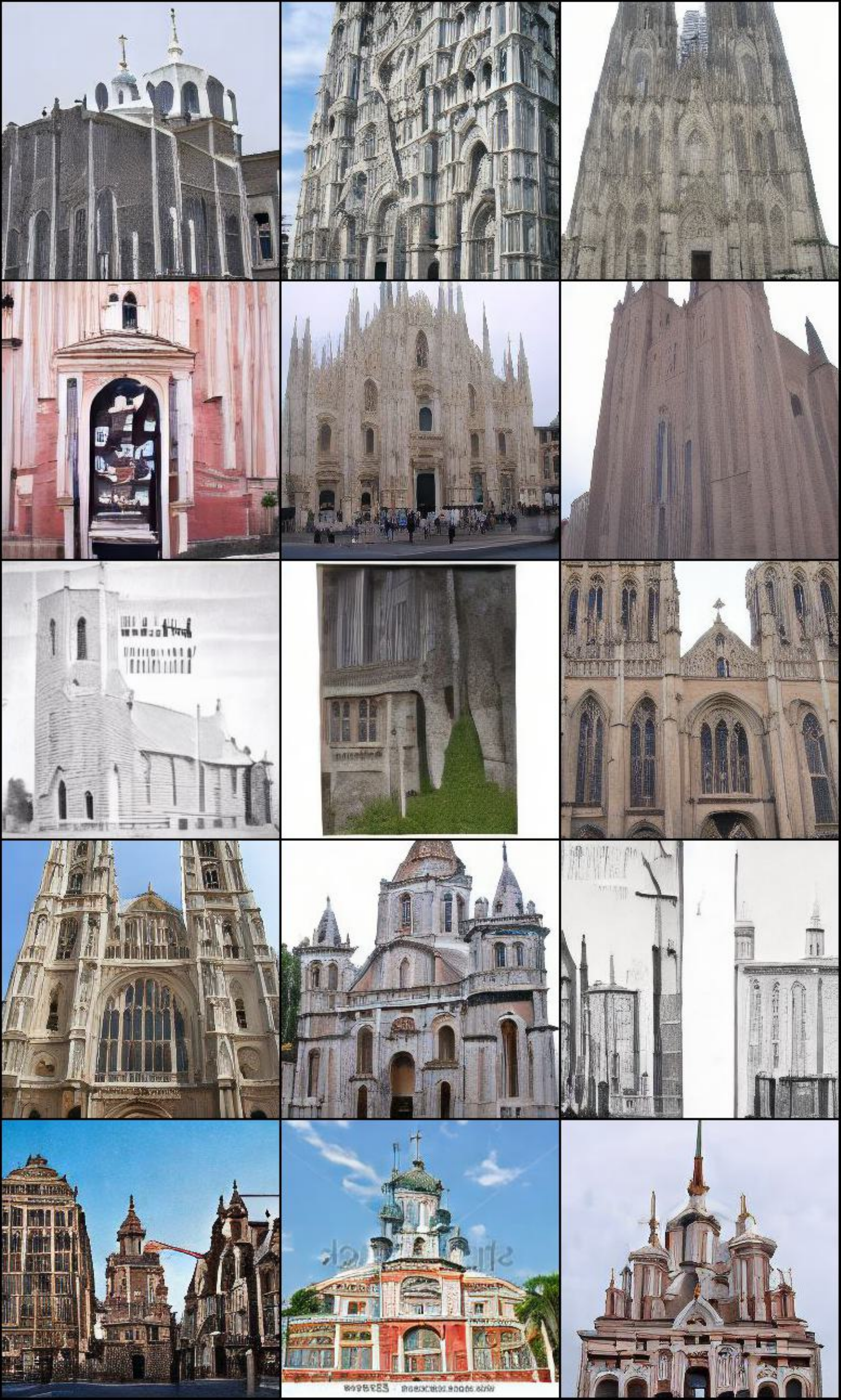}
    \vspace{0.3cm}
    \small Linear Quant (W4A8)
  \end{minipage}
  \vspace{-0.2cm}
  \caption{Random samples from our INT4 quantized 256 $\times 256$ LSUN-Church models with a fixed 
 random seed.}
  \end{center}
\end{figure*}

\begin{figure*}[h]
  \begin{center}
  \vspace{0.2cm}
  \begin{minipage}[c]{0.24\linewidth}
    \centering
    \includegraphics[width=0.99\textwidth]{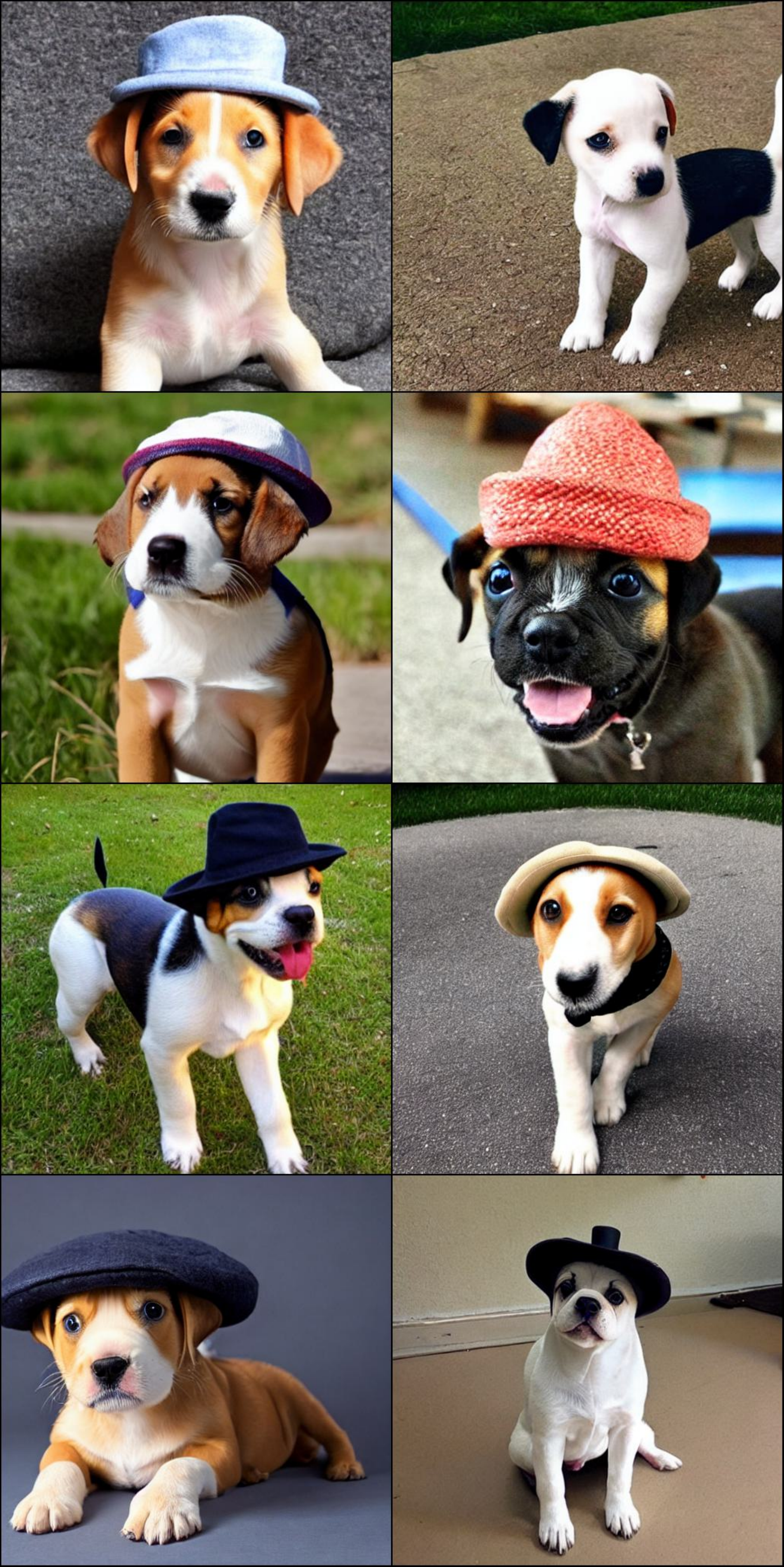}
    \vspace{0.3cm}
    \small Full Precision
  \end{minipage}\hfill
  \begin{minipage}[c]{0.24\linewidth}
    \centering
    \includegraphics[width=0.99\textwidth]{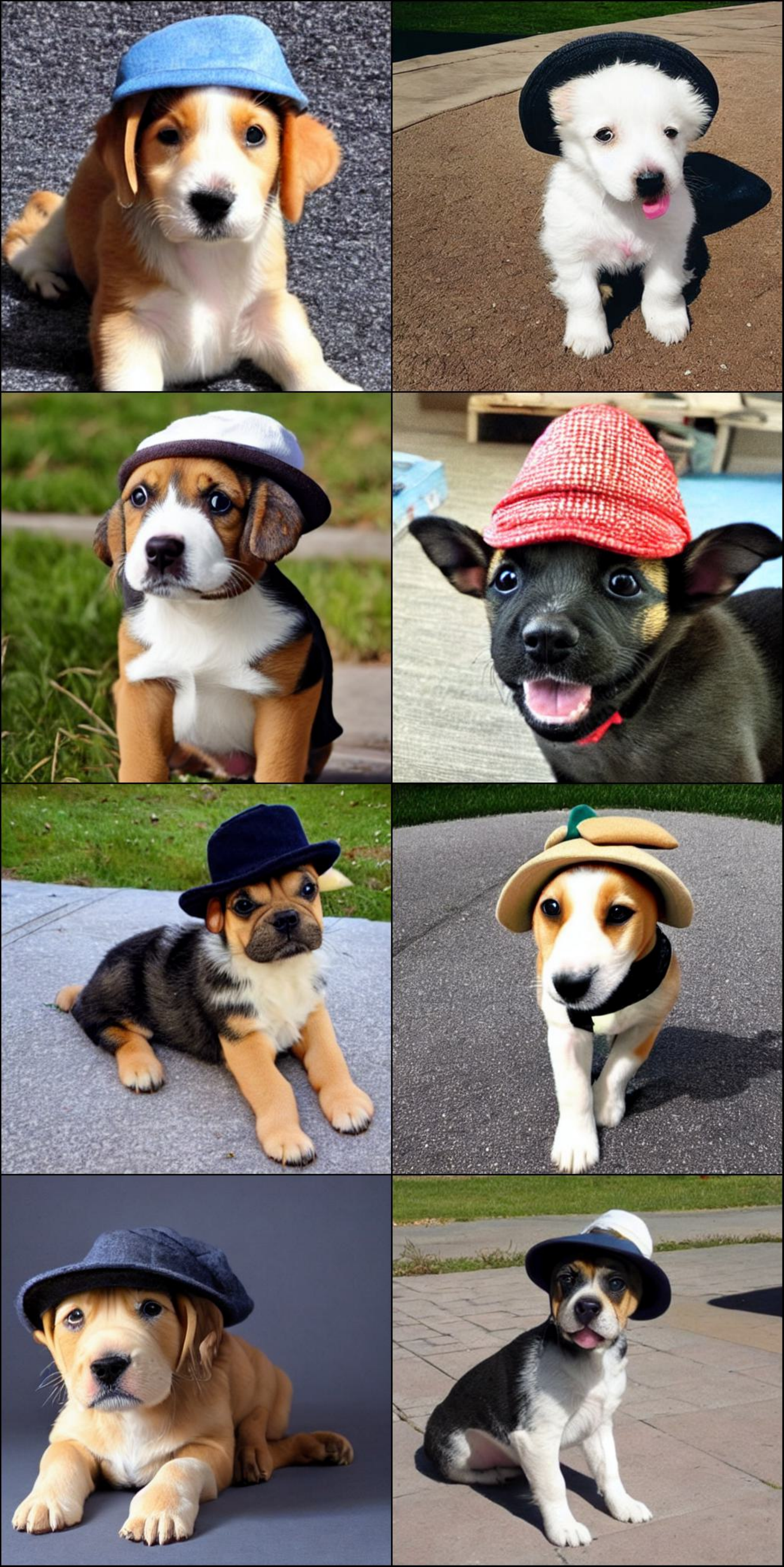}
    \vspace{0.3cm}
    \small \name (W4A32)
  \end{minipage}\hfill
  \begin{minipage}[c]{0.24\linewidth}
    \centering
    \includegraphics[width=0.99\textwidth]{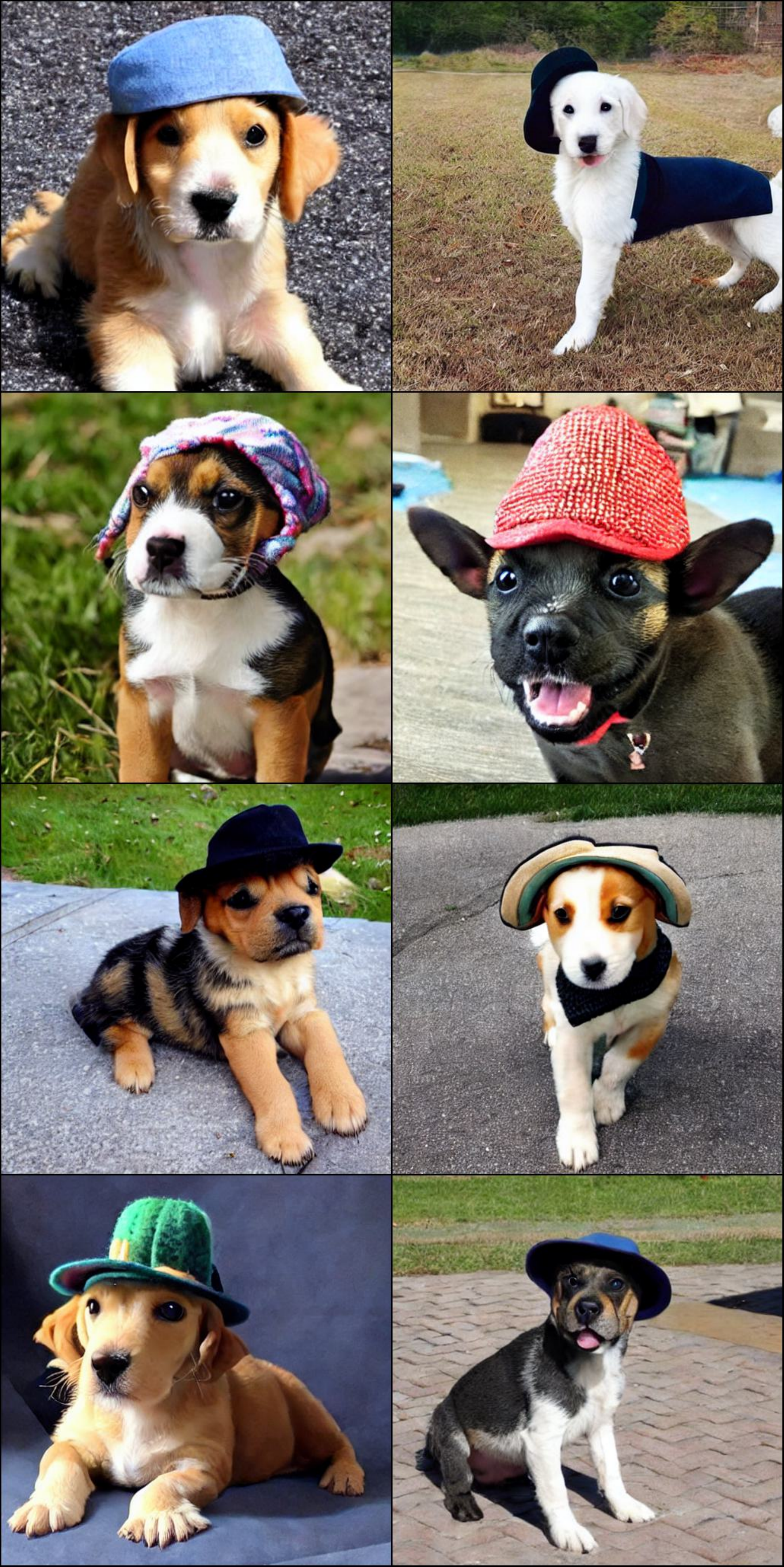}
    \vspace{0.3cm}
    \small \name (W4A8)
  \end{minipage}\hfill
  \begin{minipage}[c]{0.24\linewidth}
    \centering
    \includegraphics[width=0.99\textwidth]{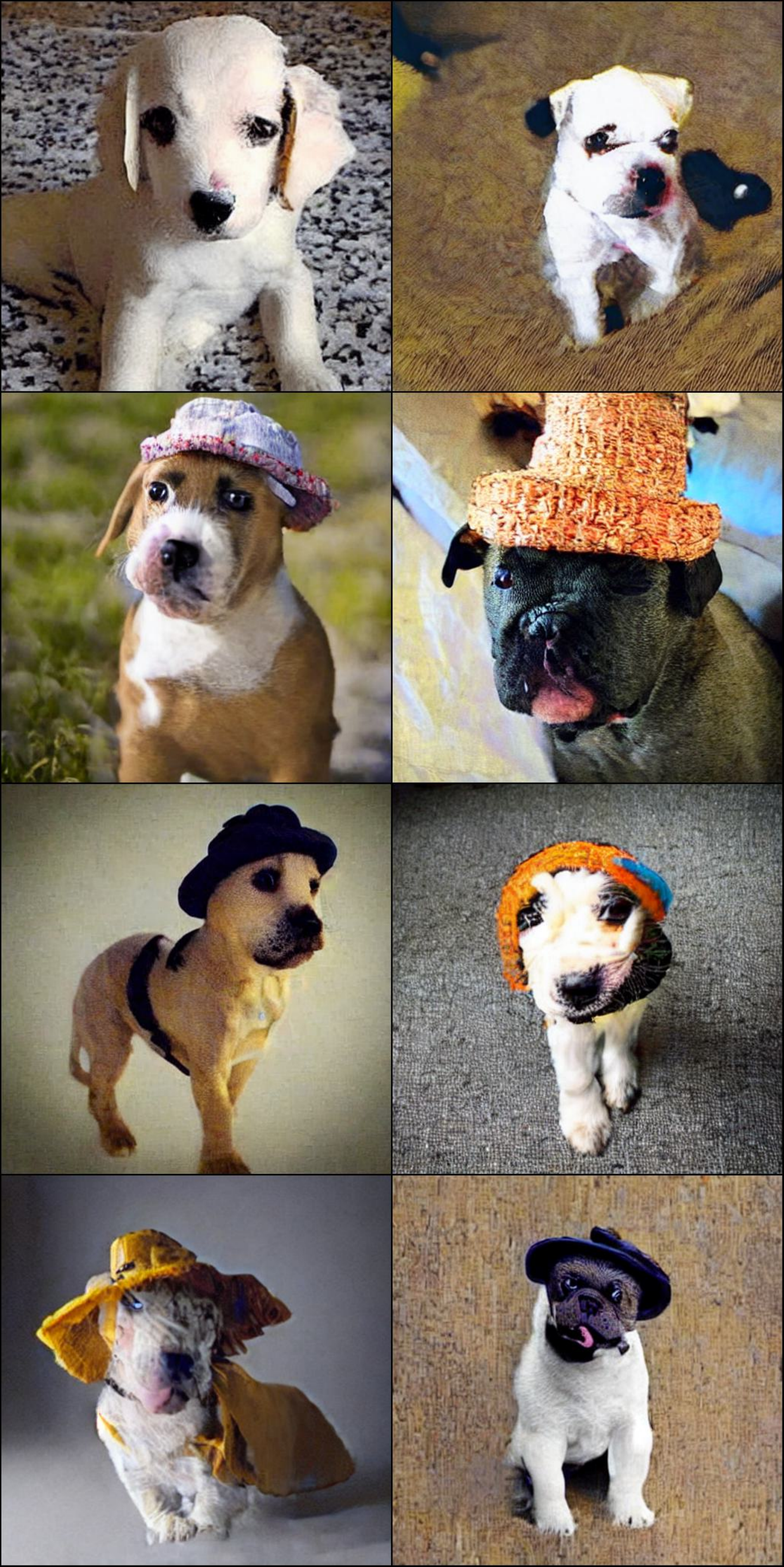}
    \vspace{0.3cm}
    \small Linear Quant (W4A32)
  \end{minipage}\hfill
  Prompt: \textit{“A puppy wearing a hat.”} \\
  \vspace{0.4cm}
  \begin{minipage}[c]{0.24\linewidth}
    \centering
    \includegraphics[width=0.99\textwidth]{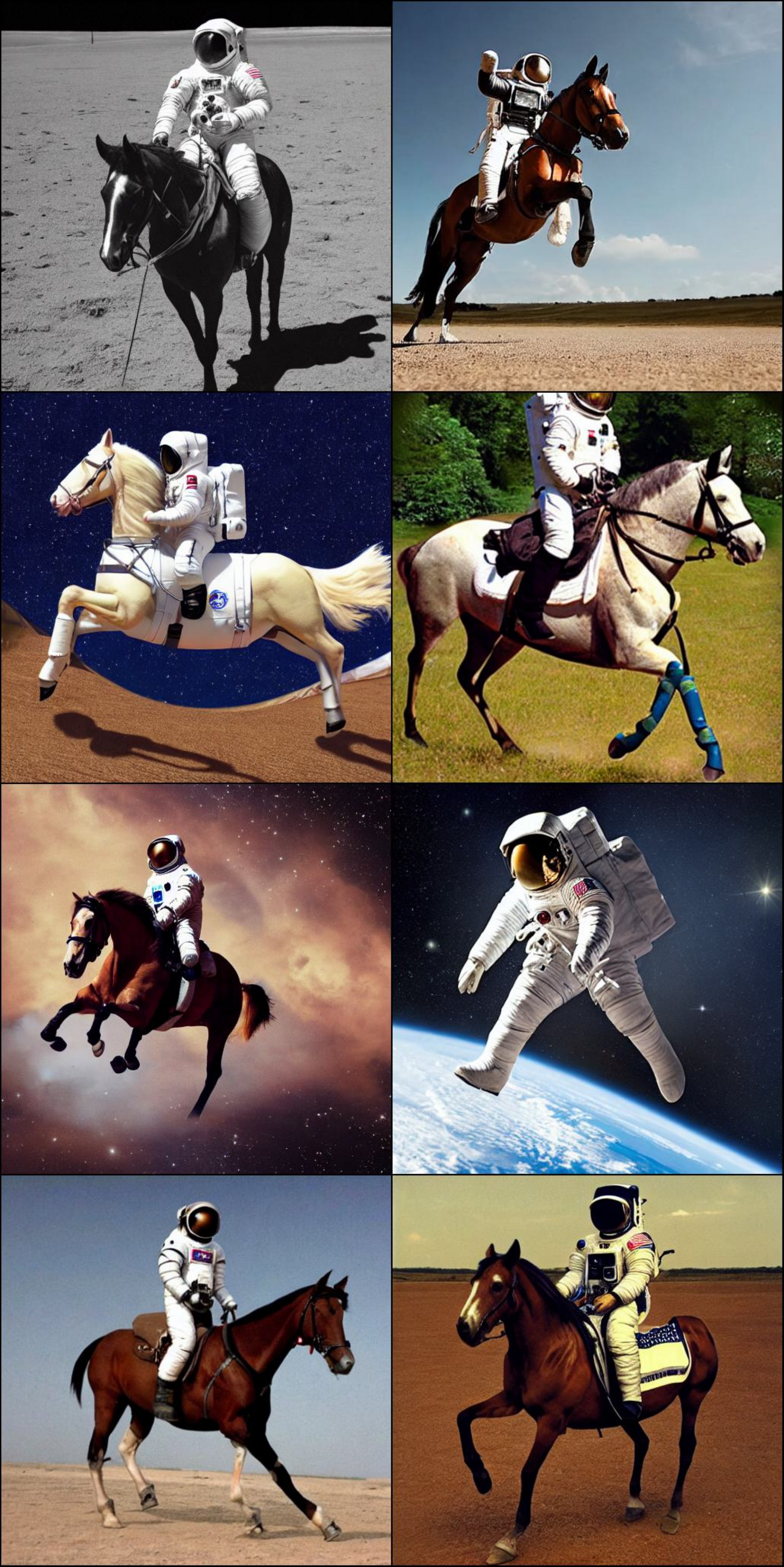}
    \vspace{0.3cm}
    \small Full Precision
  \end{minipage}\hfill
  \begin{minipage}[c]{0.24\linewidth}
    \centering
    \includegraphics[width=0.99\textwidth]{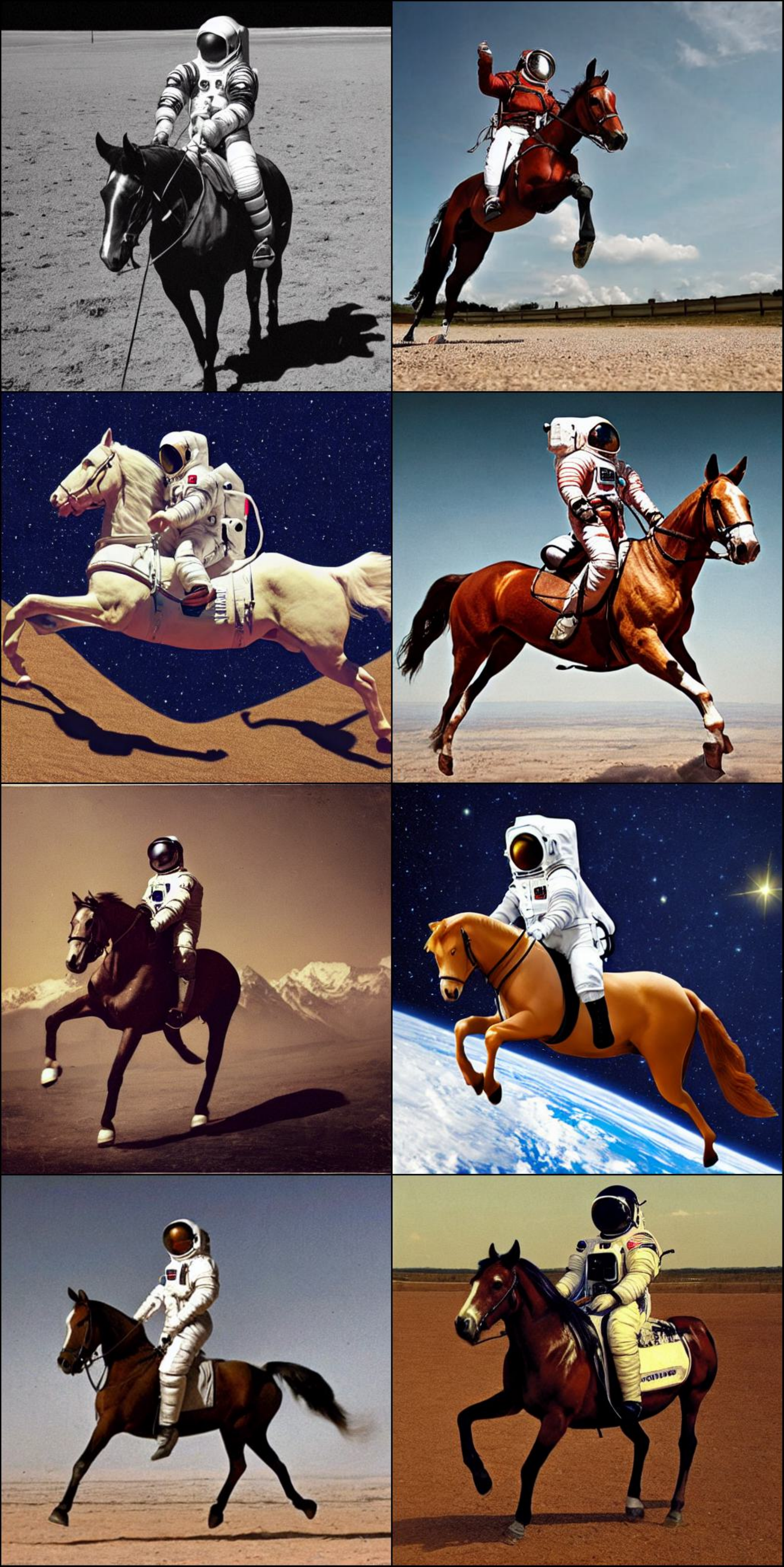}
    \vspace{0.3cm}
    \small \name (W4A32)
  \end{minipage}\hfill
  \begin{minipage}[c]{0.24\linewidth}
    \centering
    \includegraphics[width=0.99\textwidth]{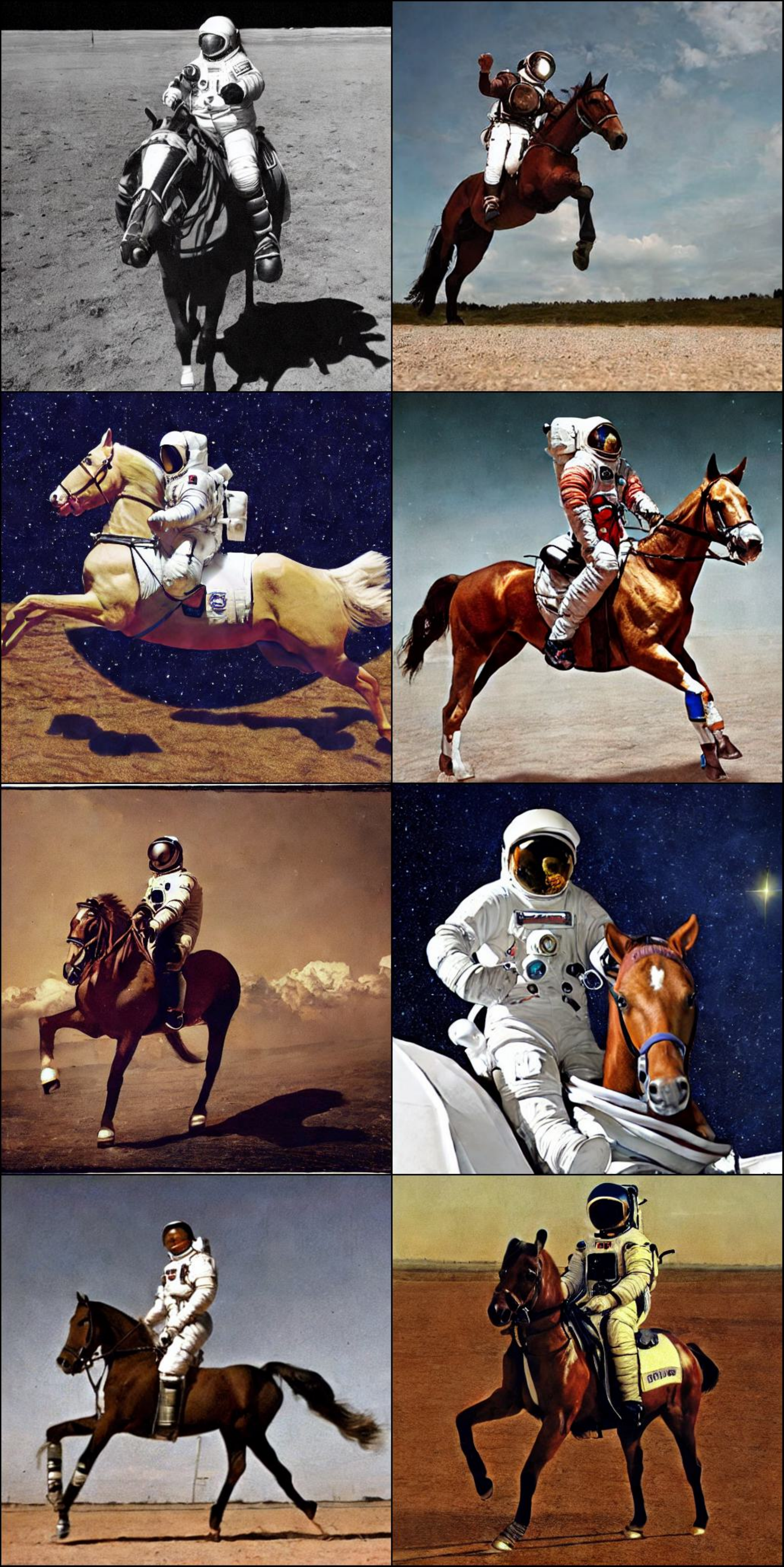}
    \vspace{0.3cm}
    \small \name (W4A8)
  \end{minipage}\hfill
  \begin{minipage}[c]{0.24\linewidth}
    \centering
    \includegraphics[width=0.99\textwidth]{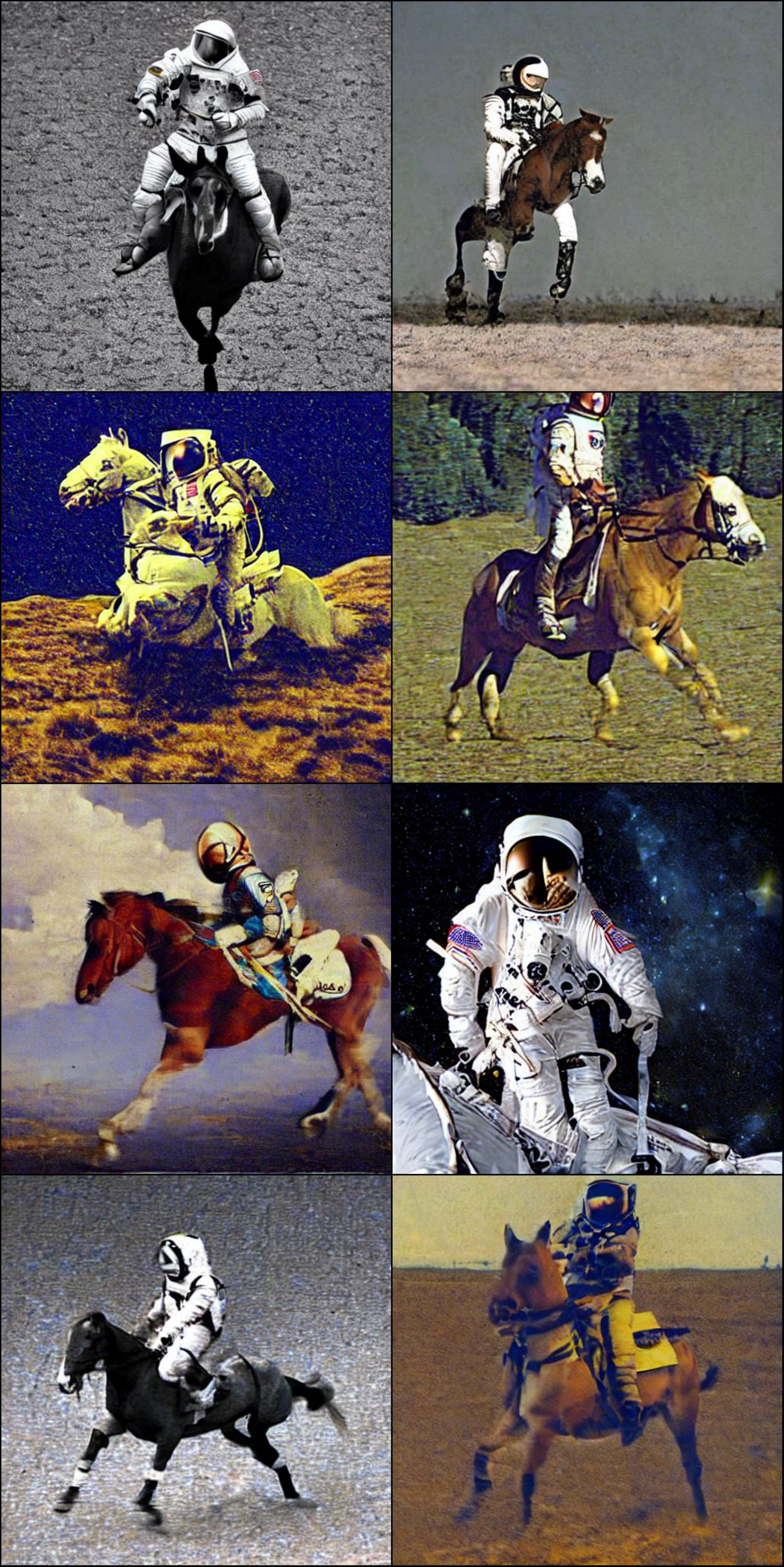}
    \vspace{0.3cm}
    \small Linear Quant (W4A32)
  \end{minipage}
  Prompt: \textit{“A photograph of an astronaut riding a horse.”} \\
  \caption{Text-guided image generation on 512 $\times$ 512 LAION-5B from our INT4 quantized \sd model with a fixed random seed.}
  \label{fig:sd_results_append}
  \end{center}
\end{figure*}